# Physics-informed radial basis network (PIRBN): A local approximating neural network for solving nonlinear partial differential equations

Jinshuai Bai[a,b,f], Gui-Rong Liu[c,*], Ashish Gupta[b,d,e], Laith Alzubaidi[a,b], Xi-Qiao Feng[f], YuanTong Gu[a,b,*]

[a] School of Mechanical, Medical and Process Engineering, Queensland University of Technology, Brisbane, QLD 4000, Australia

[b] ARC Industrial Transformation Training Centre—Joint Biomechanics, Queensland University of Technology, Brisbane, QLD 4000, Australia

[c] Department of Aerospace Engineering and Engineering Mechanics, University of Cincinnati, OH 45221, USA

[d] Queensland Unit for Advanced Shoulder Research, Brisbane, QLD, 4000, Australia

[e] Greenslopes Private Hospital, Brisbane, QLD, 4120, Australia

[f] Institute of Biomechanics and Medical Engineering, AML, Department of Engineering Mechanics, Tsinghua University, Beijing 100084, China

* Corresponding authors.

E-mail addresses: yuantong.gu@qut.edu.au (Y. Gu), liugr@ucmail.uc.edu (G-R. Liu)

## Abstract

Our recent intensive study has found that physics-informed neural networks (PINN) tend to be local approximators after training. This observation led to the development of a novel physics-informed radial basis network (PIRBN), which is capable of maintaining the local approximating property throughout the entire training process. Unlike deep neural networks, a PIRBN comprises only one hidden layer and a radial basis "activation" function. Under appropriate conditions, we demonstrated that the training of PIRBNs using gradient descendent methods can converge to Gaussian processes. Besides, we studied the training dynamics of PIRBN via the neural tangent kernel (NTK) theory. In addition, comprehensive investigations regarding the initialisation strategies of PIRBN were conducted. Based on numerical examples, PIRBN has been demonstrated to be more effective than PINN in solving nonlinear partial differential equations with high-frequency features and ill-posed computational domains.

Moreover, the existing PINN numerical techniques, such as adaptive learning, decomposition and different types of loss functions, are applicable to PIRBN. The programs that can regenerate all numerical results are available at https://github.com/JinshuaiBai/PIRBN.

*Keywords: Physics-informed Neural Network, physics-informed radial basis network, partial differential equations, neural tangent kernel,*

## 1. Introduction

Deep learning (DL) techniques have been extensively used to solve partial differential equations (PDEs) over the past decade [1]. Among these, the physics-informed neural network (PINN) has earned increasing interest because it can be trained using both observation data (i.e. boundary conditions) and PDEs themselves [2]. By training neural networks with PDEs, PINN predictions show remarkable accuracy and magnificent generalisation property [3]. Now, PINN has been widely applied to various systems that are governed by PDEs, including mechanics modelling [4-15], material exploration [16-23], and medical studies [24-34], to name but a few.

Despite its empirical success, PINN suffers from difficulties in solving PDEs with high-frequency features or have ill-posed computational domains [2, 35], and thus great efforts have been witnessed in improving PINN. Choosing the appropriate physics laws to design effective physics-informed loss functions is one way to deal with the challenges[36]. Samaniego et al [4] and Kharazmi et al. [37] formulated physics-informed loss functions for PINN training using the variational method. Haghighat et al. [38] and Yu et al. [39] embed not only PDEs but also gradient information obtained from sample points in the physics-informed loss function, rendering a significant enhancement in the performance of PINN for predicting gradient fields. Fuhg et al. [40] composed a mixed-form physics-informed loss function by integrating the variational formulation and PDEs. Bai et al. [41] integrated the weighted residual method to weaken the constraints from PDEs. Apart from the above, decomposition and transformation with regard to the computational domain is another way to address the challenges. Jagtap et al. [42] and Kharazmi et al. [37] divided computational domains into pieces. In their work, the solution of PDEs in each subdomain was predicted by a corresponding PINN. Dong and Li [43] proposed the local extreme learning framework that can ensure the continuity conditions among the decomposed domains. Based on the decomposition techniques, Shukla et al. [44] proposed a parallel computing technique to improve the training efficiency of PINNs. Nguyen-Thanh et al. [45] applied the isoparametric technique to pre-normalise complex computational domains into simple geometries.

Recent advances in terms of the theoretical study of DL, the so-called neural tangent kernel (NTK) theory, have also shed light to PINN. The original NTK theory was proposed by Jacot et al. [46] and proved that training of infinite width neural networks with enough small learning rate can converge to Gaussian processes [47, 48], which provided insights into DL training dynamics from the theoretical regard [49]. The NTK theory was then introduced to PINN by Wang et al. [50]. In their work, the training of PINN was analysed via the lens of NTK theory. More importantly, an NTK-based adaptive learning scheme was proposed to balance the loss terms automatically. With the help of the NTK-based adaptive learning scheme, different physics in PINNs can be equally trained and the PINNs can more effectively deal with problems that exhibit high-frequency features. Later, Wang et al. [51, 52] reported the connection between the NTK and the spectral bias in PINN, and tuned neural network structures to address multi-scale problems and long-term evolution problems. Li et al. [53] analysed the error estimation of PINNs through the NTK theory. Xu et al. [54] studied the training dynamics of PINNs for inverse problems via the NTK theory and proposed a novel gradient descendent algorithm.

Inspired by the aforementioned works, this work starts by studying the training dynamics of PINN via the NTK theory. By visualising the evolution of a PINN's NTK throughout training, we found that a PINN tends to be a local approximator after training. Besides, when coping with PDEs that exhibit high-frequency features or have ill-posed computational domains, PINN fails for predictions due to the difficulties of training PINN into a local approximator. Thus, based on this finding, we tailored a neural network structure, namely the physics-informed radial basis network (PIRBN), that naturally possesses the local approximation property. The main contributions of this work are summarised as follows:

i) From numerical experiments and the NTK theory, we found that PINNs tend to become local approximators during training via gradient descendent algorithms.

ii) Based on the above finding, we proposed the PIRBN, which has the local approximation property throughout the training.

iii) We prove that training the proposed PIRBN via gradient descendent algorithms also converges to Gaussian Processes when the width of the PIRBN tends to be infinite with an infinitesimal small learning rate.

iv) Comprehensive studies of the proposed PIRBN regarding parameter initialisation strategies, sizes of sample points and selections of radial basis activation functions are provided.

v) Challenging PDEs examples, which possess high-frequency features and ill-posed computational domains, are presented to show the performance of PIRBN with respect to PINN. Those problems that require domain decomposition techniques and multiple-layer PINNs can be now effectively solved by using single-layer PIRBNs.

The paper is organised as follows: In Section 2, the basic conceptions of PINN and the NTK theory for PINN are briefly recalled. Meanwhile, by visualising the NTK of PINN through training processes, the local approximation property of PINN is observed. In Section 3, based on the finding, the PIRBN is proposed and formulated. Besides, the training dynamics of PIRBN are analysed by the NTK theory and the initialisation strategies of PIRBN are discussed in detail. In Section 4, several numerical examples, which are challenged for PINN, are conducted to show the performance of PIRBN. In Section 5, the conclusions of this work are summarised, and future perspectives are provided.

## 2. A recap of PINN and NTK theory for PINN

### 2.1. Physics-informed neural network

We first recap the basic conceptions of the initial PINN [55]. Consider a set of PDEs with the boundary condition as

$$\begin{aligned} G[u] - g(\mathbf{x}^g) = 0, \quad \text{for } \mathbf{x}^g \in \Omega, \\ B[u] - b(\mathbf{x}^b) = 0, \quad \text{for } \mathbf{x}^b \in \partial\Omega, \end{aligned} \quad (1)$$

where $G[\cdot]$ is prescribed partial differential operators and $B[\cdot]$ is the boundary condition operator, $\Omega$ denotes the computational domain in $\mathbb{R}^n$, and $u(x): \mathbb{R}^n \to \mathbb{R}$ is the solution function governed by the PDE. To solve the PDE problem, a feedforward neural network (FNN) [11] is used to approximate the solution function as

$$u(\mathbf{x}) = F(\mathbf{x}, \boldsymbol{\theta}) \quad (2)$$

where $\boldsymbol{\theta}$ represents all the trainable parameters of the FNN. To train the FNN with physics, the physics-informed loss function is formulated as

$$\mathcal{L}(\boldsymbol{\theta}) = \mathcal{L}_g(\boldsymbol{\theta}) + \mathcal{L}_b(\boldsymbol{\theta}), \quad (3)$$

where $\mathcal{L}_g$ and $\mathcal{L}_b$ are the loss terms from the PDE and the boundary condition, respectively. They can be calculated by

$$\mathcal{L}_g(\boldsymbol{\theta}) = \frac{1}{2}\sum_i^{n_g} \left| G[u(\mathbf{x}_i^g)] - g(\mathbf{x}_i^g) \right|^2,$$

$$\mathcal{L}_b(\boldsymbol{\theta}) = \frac{1}{2}\sum_i^{n_b} \left| B[u(\mathbf{x}_i^b)] - b(\mathbf{x}_i^b) \right|^2, \tag{4}$$

where $n_g$ and $n_b$ are the numbers of in-domain sample points and boundary sample points, respectively. More details regarding PINN can be found in [55].

*2.2. Neural tangent kernel theory for physics-informed neural network*

The neural tangent kernel (NTK) theory for PINN also is briefly recalled. Consider an infinite width PINN for solving a PDE shown as Eq.(1). A gradient descendent algorithm is selected as the training approach and all parameters in the network are initialised via the LeCun initialisation scheme [56]. In this manner, according to Wang et al. [50], the training process for such PINN can converge to independent and identically distributed (i.i.d.) centred Gaussian processes ($\mathcal{GP}$)

$$\lim_{d\to\infty} G\left[u\left(\mathbf{x};\boldsymbol{\theta}(t)\right)\right] \to \mathcal{GP}(0, \Sigma_g(\mathbf{x}, \mathbf{x}')),$$

$$\lim_{d\to\infty} B\left[u\left(\mathbf{x};\boldsymbol{\theta}(t)\right)\right] \to \mathcal{GP}(0, \Sigma_b(\mathbf{x}, \mathbf{x}')), \tag{5}$$

where $d$ denotes the width (number of neurons in each hidden layer) of PINN, $t$ denotes the training time [50], $\Sigma_g(\mathbf{x}, \mathbf{x}')$ and $\Sigma_b(\mathbf{x}, \mathbf{x}')$ are the covariances for corresponding Gaussian processes. Specifically, the covariance denotes the correlation of the values of target functions (i.e. $G[u]$ and $B[u]$) when the inputs are $\mathbf{x}$ and $\mathbf{x}'$. Based on that, the NTK of a PINN, $\mathbf{K}(t)$, is defined as

$$\mathbf{K}(t) = \begin{bmatrix} \mathbf{K}_{gg}(t) & \mathbf{K}_{gb}(t) \\ \mathbf{K}_{bg}(t) & \mathbf{K}_{bb}(t) \end{bmatrix}, \tag{6}$$

where

$$\left(\mathbf{K}_{gg}\right)_{ij}(t) = \left\langle \frac{dG\left[u\left(\mathbf{x}_i^g;\boldsymbol{\theta}(t)\right)\right]}{d\boldsymbol{\theta}}, \frac{dG\left[u\left(\mathbf{x}_j^g;\boldsymbol{\theta}(t)\right)\right]}{d\boldsymbol{\theta}} \right\rangle,$$

$$\left(\mathbf{K}_{gb}\right)_{ij}(t) = \left\langle \frac{dG\left[u\left(\mathbf{x}_i^g;\boldsymbol{\theta}(t)\right)\right]}{d\boldsymbol{\theta}}, \frac{dB\left[u\left(\mathbf{x}_j^b;\boldsymbol{\theta}(t)\right)\right]}{d\boldsymbol{\theta}} \right\rangle, \quad (7)$$

$$\left(\mathbf{K}_{bb}\right)_{ij}(t) = \left\langle \frac{dB\left[u\left(\mathbf{x}_i^b;\boldsymbol{\theta}(t)\right)\right]}{d\boldsymbol{\theta}}, \frac{dB\left[u\left(\mathbf{x}_j^b;\boldsymbol{\theta}(t)\right)\right]}{d\boldsymbol{\theta}} \right\rangle.$$

Using the NTK, training a PINN through gradient descendent algorithms can be written as

$$\begin{bmatrix} \dfrac{dG\left[u\left(\mathbf{x}^g;\boldsymbol{\theta}(t)\right)\right]}{dt} \\ \dfrac{dB\left[u\left(\mathbf{x}^b;\boldsymbol{\theta}(t)\right)\right]}{dt} \end{bmatrix} = -\mathbf{K}(t) \begin{bmatrix} G\left[u\left(\mathbf{x}^g;\boldsymbol{\theta}(t)\right)\right] - g(\mathbf{x}^g) \\ B\left[u\left(\mathbf{x}^b;\boldsymbol{\theta}(t)\right)\right] - b(\mathbf{x}^b) \end{bmatrix}. \quad (8)$$

As demonstrated by Wang et al. [50], the NTK of PINN with a sufficiently large width (i.e., $d \to \infty$) barely changes during the training process via gradient descendent algorithms

$$\mathbf{K}' \approx \mathbf{K}(0) \approx \mathbf{K}(t). \quad (9)$$

In this manner, training infinite width PINNs through a gradient descendent algorithm with enough small learning rate (also known as gradient flow [50]) can be regarded as solving a linear PDE problem, i.e.

$$\begin{bmatrix} G\left[u\left(\mathbf{x}^g;\boldsymbol{\theta}(t)\right)\right] \\ B\left[u\left(\mathbf{x}^b;\boldsymbol{\theta}(t)\right)\right] \end{bmatrix} \approx \left(\mathbf{I} - e^{-\mathbf{K}'t}\right) \begin{bmatrix} g(\mathbf{x}^g) \\ b(\mathbf{x}^b) \end{bmatrix}. \quad (10)$$

Now, consider the spectral decomposition of the NTK

$$\mathbf{K}' = \mathbf{Q}^T \boldsymbol{\Lambda} \mathbf{Q}, \quad (11)$$

where $Q$ is an orthogonal matrix and $\Lambda$ is a diagonal matrix of eigenvalues. By substituting Eq. (11) into (10), we obtain

$$\mathbf{Q} \begin{bmatrix} G\left[u\left(\mathbf{x}^g;\boldsymbol{\theta}(t)\right)\right] - g(\mathbf{x}^g) \\ B\left[u\left(\mathbf{x}^b;\boldsymbol{\theta}(t)\right)\right] - b(\mathbf{x}^b) \end{bmatrix} \approx - \begin{bmatrix} e^{-\lambda_1 t} & 0 & \cdots & 0 \\ 0 & e^{-\lambda_2 t} & \cdots & 0 \\ \vdots & \vdots & \ddots & \vdots \\ 0 & 0 & \cdots & e^{-\lambda_n t} \end{bmatrix} \mathbf{Q} \begin{bmatrix} g(\mathbf{x}^g) \\ b(\mathbf{x}^b) \end{bmatrix}, \quad (12)$$

where $\lambda_i$ denotes the $i^{\text{th}}$ eigenvalue and is corresponding to the convergence rate of $\mathbf{Q}\begin{bmatrix} G[u(\mathbf{x}^g;\boldsymbol{\theta}(t))] - g(\mathbf{x}^g) \\ B[u(\mathbf{x}^b;\boldsymbol{\theta}(t))] - b(\mathbf{x}^b) \end{bmatrix}$. Details regarding the proof can be found in [50].

With the aid of NTK and the corresponding NTK-based adaptive training method [50], the performance of PINN can be significantly improved for problems with high-frequency features. For example, consider a PDE and its boundary conditions as

$$\frac{d^2}{dx^2}u(x) - 4\mu^2\pi^2 \sin(2\mu\pi x) = 0, \quad \text{for } x \in [0,1],$$
$$u(0) = u(1) = 0,$$
(13)

where $\mu$ is a constant that controls the frequency feature of the PDE solution. The analytical solution is given as

$$u(x) = \sin(2\mu\pi x). \tag{14}$$

**Fig. 1** shows the predictions from PINN (one hidden layer with 61 neurons) via the NTK adaptive training methods when $\mu = 4$. Eq. (3) is used as the loss function for PINN. The *tanh* function is selected as the activation function and all the parameters in the PINN are initialised via the LuCun initialisation [46]. 51 sample points are uniformly distributed in the computational domain. The neural network is trained for $10^4$ iterations with the Adam optimiser (learning rate 0.001). As shown in the figure, the PINN performs well for this problem and the maximum point-wise error is $1.91 \times 10^{-3}$.

Additionally, it is worth highlighting that the normalised NTK matrix, $\mathbf{K}_g$, evolutes to a diagonal matrix during the training process, as shown in **Fig. 1**(d)-(f). This suggests that the neural network is trained to be a local approximator; that is, changing trainable parameters to approach a given data's ground truth is less likely to affect other data. The same diagonal pathology during training is observed from higher frequency cases where $\mu = 8$, as shown in **Fig. 2**(d)-(f). We also note that PINNs with multiple hidden layers FNNs also have this local approximation property. The details regarding that are provided in Appendix A. Nevertheless, for the higher frequency case, the FNN produce poor predictions compared to the ground truth, as shown in **Fig. 2**(a)(b), suggesting that a single-layer FNN with only 61 neurons is not sufficient enough for solving this high-frequency PDE problem. In order to achieve higher accuracy, a multiple hidden layers FNN is normally applied. This is because, intuitively, an FNN with an increasing number of hidden layers is more powerful in approximating a continuous function [57].

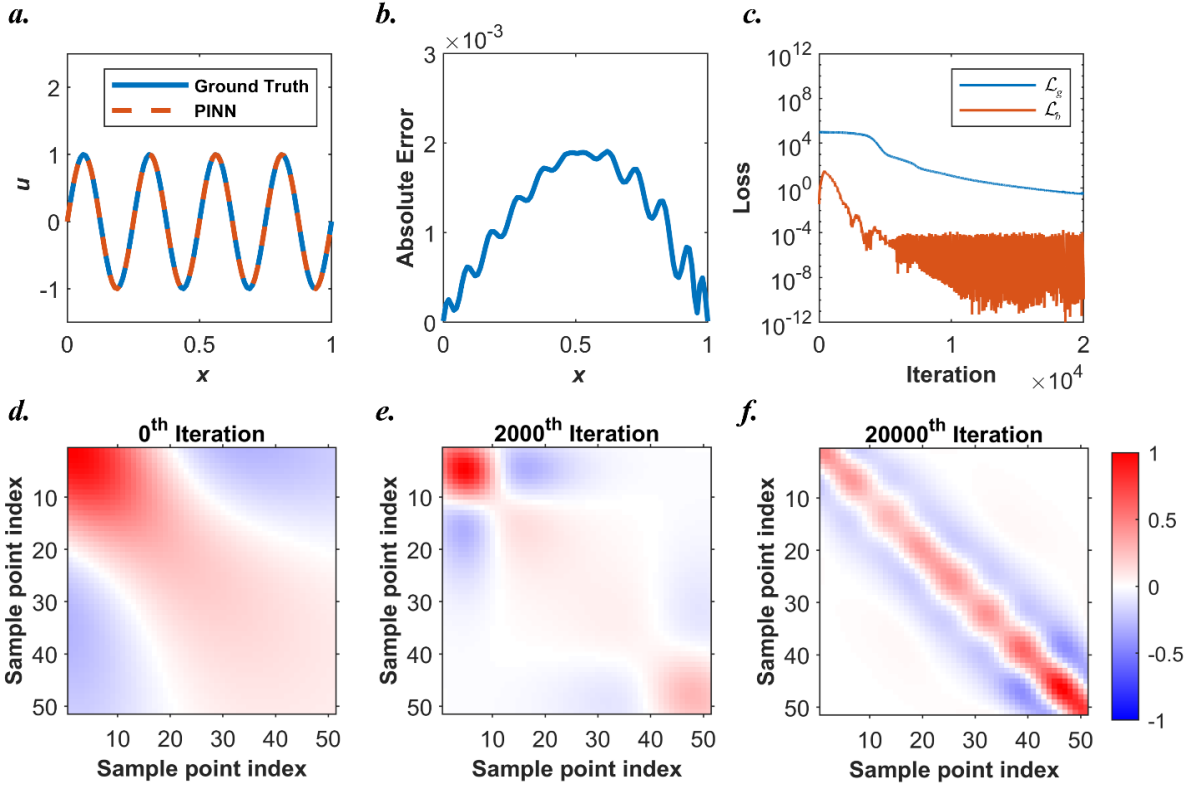

**Fig. 1.** Results from a PINN (single hidden layer with 61 neurons per layer) for solving Eq. (13) when $\mu = 4$. (a) Comparisons between the PINN predictions and the ground truth. (b) Point-wise absolute error plot. (c) Loss history of the PINN during the training process. (d) The normalised $\mathbf{K}_g$ at $0^{th}$ iteration. (e) The normalised $\mathbf{K}_g$ at $2000^{th}$ iteration. (f) The normalised $\mathbf{K}_g$ at $20000^{th}$ iteration.

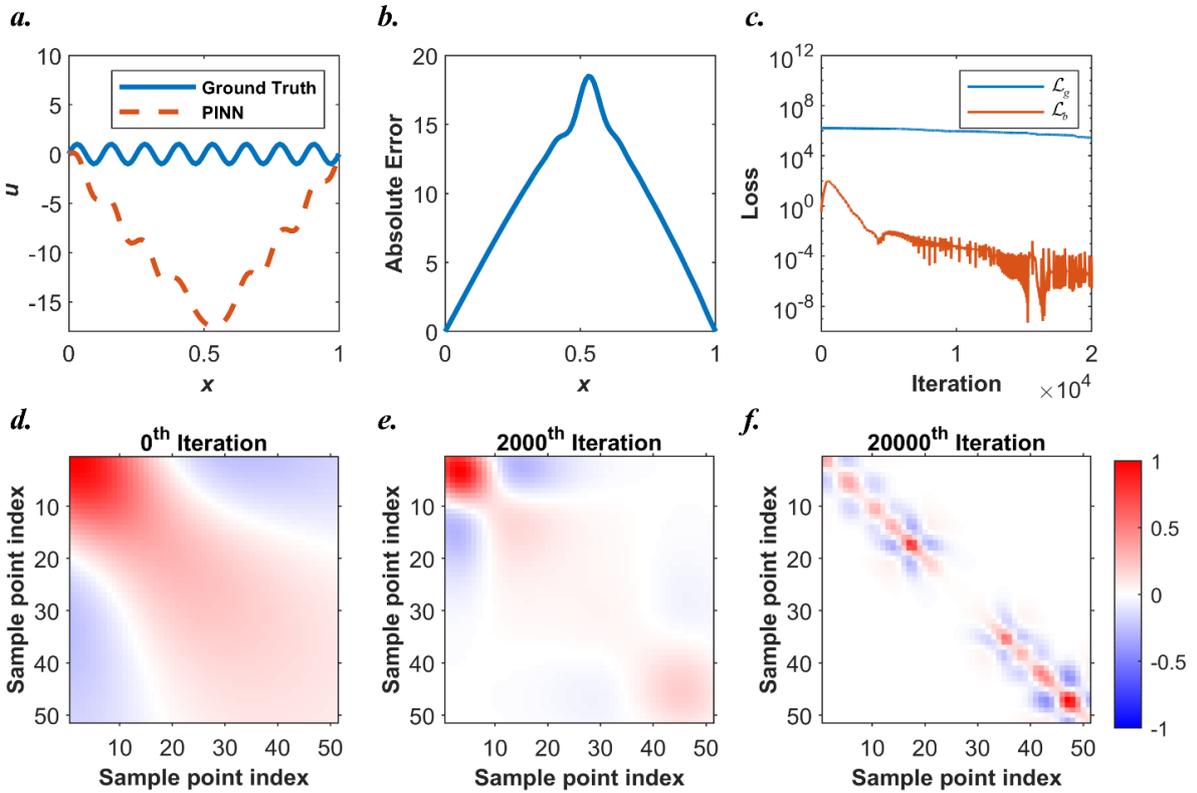

**Fig. 2.** Results from a PINN (single hidden layer with 61 neurons per layer) for solving Eq. (13) when $\mu = 8$. (a) Comparisons between the PINN predictions and the ground truth. (b) Point-wise absolute error plot. (c) Loss history of the PINN during the training process. (d) The normalised $\mathbf{K}_g$ at $0^{th}$ iteration. (e) The normalised $\mathbf{K}_g$ at $2000^{th}$ iteration. (f) The normalised $\mathbf{K}_g$ at $20000^{th}$ iteration.

Besides, PINN also struggles to cope with problems when problem domains are ill-posed or not normalised. For example, consider Eq. (13) is translated to right by 100 to:

$$\frac{d^2}{dx^2}u(x-100) - 4\mu^2\pi^2 \sin(2\mu\pi(x-100)) = 0, \quad \text{for } x \in [100,101], \quad (15)$$
$$u(100) = u(101) = 0.$$

Correspondingly, the problem domain is changed from $x \in [0,1]$ to $x \in [100,101]$. The PINN with the previous initialisation can suffer from severe failure, as shown in **Fig. 3**. It is found that all the elements of the normalised NTK of the PINN stay close to 1 from the beginning, suggesting that the sample points are highly correlated during the training process. In another word, slightly changing the trainable parameters inside the PINN can equally affect the predictions for all the sample points, which brings significant difficulties for the gradient descendent algorithm to minimise the loss. This is further demonstrated by the training plot shown in **Fig. 3**(c), where the loss term from the PDE, $\mathcal{L}_g$, almost remains at $10^4$ and barely decreases along with the training.

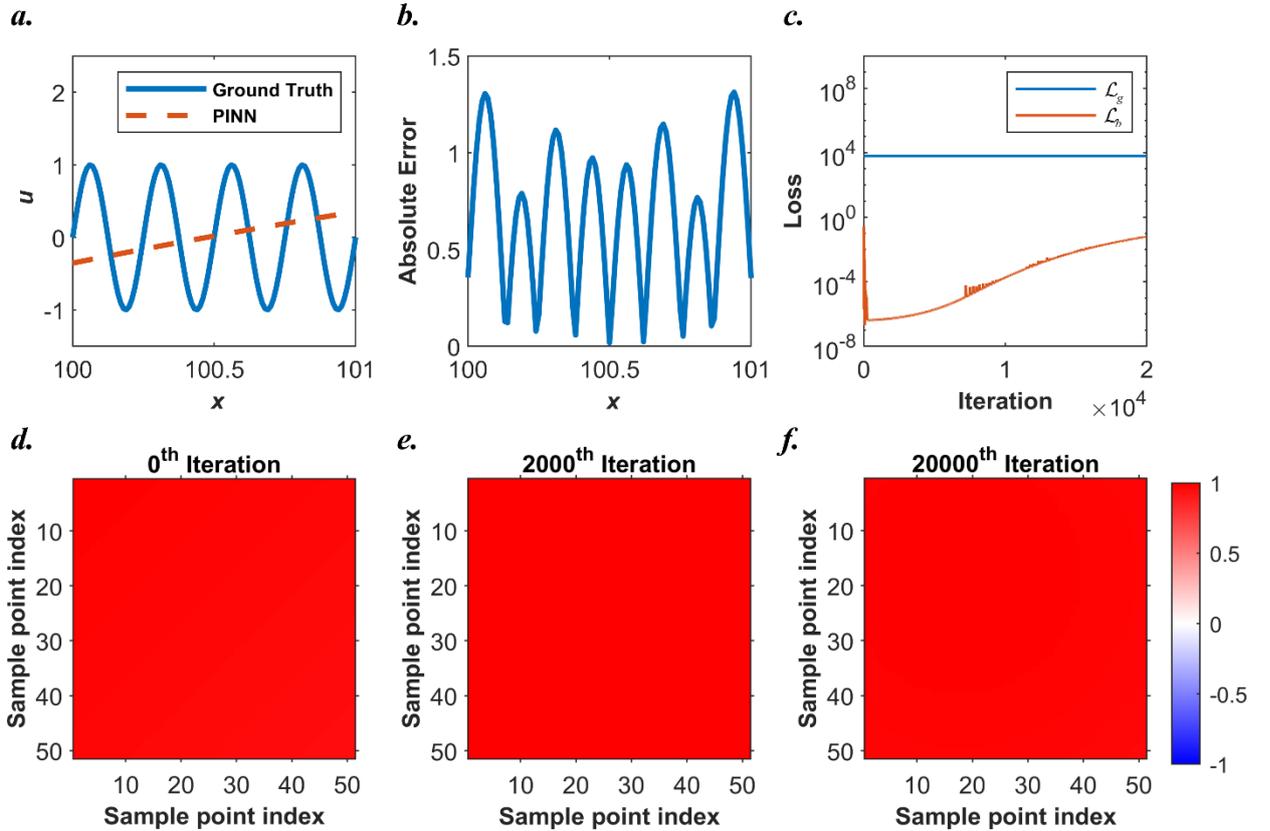

**Fig. 3.** Results from a PINN (single hidden layer with 61 neurons per layer) for solving Eq. (15) when $\mu = 4$. (a) Comparisons between the PINN predictions and the ground truth. (b) Point-wise absolute error plot. (c) Loss history of the PINN during the training process. (d) The normalised $\mathbf{K}_g$ at $0^{th}$ iteration. (e) The normalised $\mathbf{K}_g$ at $2000^{th}$ iteration. (f) The normalised $\mathbf{K}_g$ at $20000^{th}$ iteration.

## 3. Physics-informed radial basis network

As reported in Section 2, PINNs tend to be trained to a local approximator during training. For those PINNs that fail to achieve local approximation property, they can be hardly trained by physics-informed loss functions. In this manner, a question raises: will it helps if a local neural network is applied? In other words, will it be more effective and efficient to train neural networks if such networks already satisfy the local approximation property before training? To answer the questions, we propose the PIRBN and test its performance.

The original radial basis network is a single-layer neural network proposed by Broomhead and Lowe [58]. In the original radial basis network, the radial basis function (RBF), $\vartheta$: $\mathbb{R}^n \to \mathbb{R}$, is used as the activation function. The value of an RBF depends only on the distance from a given centre, $\mathbf{c}$, to the input $\mathbf{x}$

$$\vartheta(\mathbf{x}) = f(\|\mathbf{x}-\mathbf{c}\|), \tag{16}$$

For example, the most prevailingly used RBF in the radial basis network is the Gaussian function which reads

$$f(\|\mathbf{x}-\mathbf{c}\|) = e^{-b^2\|\mathbf{x}-\mathbf{c}\|^2}, \tag{17}$$

where $b$ is a variable that controls the shape of the RBF, as shown in **Fig. 4**. Besides the Gaussian function, typical RBFs also include inverse quadratic function, inverse multiquadric function, thin plate spline function, to name but a few [59]. When using the Gaussian function as the activation function, the mapping between the input and output in a radial basis network can be mathematically expressed as

$$y = \frac{1}{\sqrt{d}} \sum_i^d a_i \vartheta_i(\mathbf{x}), \tag{18}$$

where

$$\vartheta_i(x) = e^{-b_i^2\|\mathbf{x}-\mathbf{c}_i\|^2}, \tag{19}$$

where $d$ is the width of the radial basis network, and $a_i$ and $b_i$ are trainable parameters.

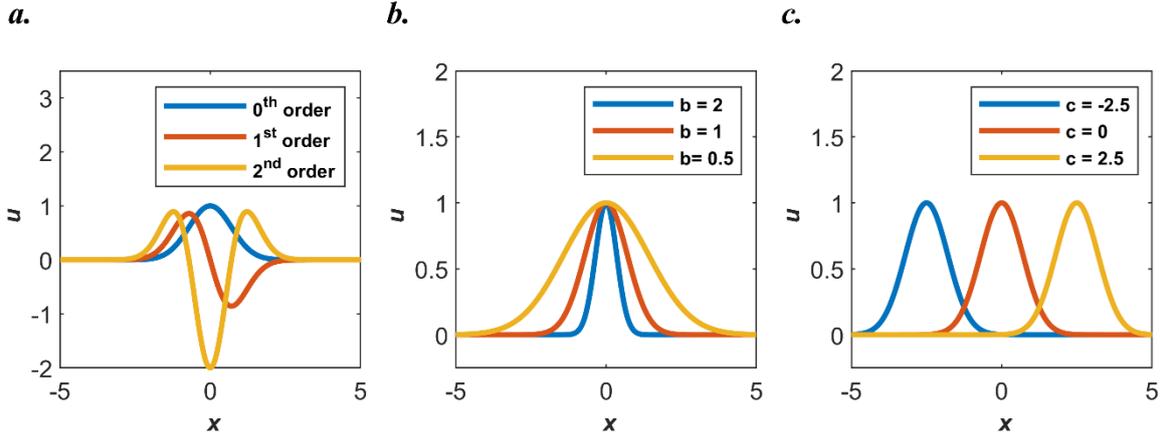

**Fig. 4.** An example of the Gaussian function. (a) $0^{th}$, $1^{st}$ and $2^{nd}$ orders derivatives of the Gaussian function. (b) The Gaussian function with different $b$. (c) The Gaussian function with different $c$.

In this work, the radial basis network is trained with corresponding physics and its loss function can be written as Eq. (3). It is worth highlighting that, the centres of all RBF neurons of the radial basis network are constants during training, which is different from the original radial basis network. For convenience in discussion, we term this kind of radial basis network as the physics-informed radial basis network (PIRBN). An example of the PIRBN is shown in **Fig. 5**.

While using PIRBNs, each RBF neuron is only activated when the input is near the centre of the neuron. Intuitively, a PIRBN exhibits the local approximation property. In what follows, the NTK theory is applied to study the training dynamics of PIRBN and therefore proves the local approximation property of PIRBN during training. Furthermore, corresponding initialisation strategies are also presented and discussed in detail.

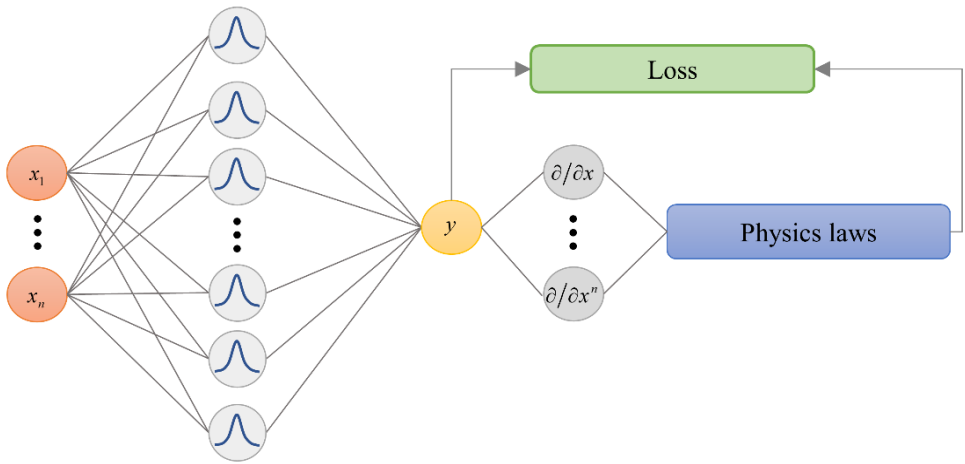

**Fig. 5.** The structure of a physics-informed radial basis network (PIRBN).

*3.1. Analysing PIRBN via the NTK theory*

As aforementioned, the NTK theory provides a powerful tool to investigate the training process of PINNs. The training process of a PIRBN via gradient descendent algorithms can also be analysed by the NTK theory, which is presented here.

As an example, consider the problem defined in Eq. (15), to which we had difficulty training a PINN earlier. A PIRBN with the Gaussian function is now constructed to solve this problem. The Adam optimiser is also used. Before analysis, the parameters $a_i$ in the PIRBN are initialised with i.i.d. random variables satisfy $\mathcal{N}(0,1)$ distribution and $c_i$ are initialised as constants so that the RBF neurons can be uniformly distributed in the computational domain. Besides, $b_i$ in the PIRBN are initialised to a given value. These are important prerequisites to obtaining the following theorems.

**Theorem 3.1.1.** *As $d \to \infty$, an initialised PIRBN predictions converge to centred Gaussian processes*

$$\begin{aligned}\lim_{d\to\infty}\frac{d^2}{dx^2}u(\mathbf{x};\boldsymbol{\theta}(t)) &\to \mathcal{GP}(0,\Sigma_g(\mathbf{x},\mathbf{x}')),\\ \lim_{d\to\infty}u(\mathbf{x};\boldsymbol{\theta}(t)) &\to \mathcal{GP}(0,\Sigma_b(\mathbf{x},\mathbf{x}')),\end{aligned} \quad (20)$$

*where $\Sigma_g(x, x')$ and $\Sigma_b(x, x')$ are their corresponding covariances*

$$\begin{aligned}\Sigma_g(\mathbf{x},\mathbf{x}') &= \mathbb{E}\left[a^2\ddot{\vartheta}(\mathbf{x})\ddot{\vartheta}(\mathbf{x}')\right],\\ \Sigma_b(\mathbf{x},\mathbf{x}') &= \mathbb{E}\left[\vartheta(\mathbf{x})\vartheta(\mathbf{x}')\right].\end{aligned} \quad (21)$$

*Proof.* The proof of this theorem is given in Appendix B.1.

**Remark 3.1.2.** Theorem 3.1.1 indicates that, as $d \to \infty$, the training process of a PIRBN can be regarded as Gaussian processes [46]. In another word, training a PIRBN via gradient descendent algorithms can also be regarded as a kernel-based regression problem [60]. Therefore, based on Theorem 3.1.1 and followed by the definition of the NTK for a PINN, we can also define the NTK of a PIRBN as Eq. (6) and Eq. (7). It is worth noting that, the NTK of a PIRBN is the covariance matrix for the Gaussian processes. That is, each element in the NTK, $K_{ij}$, indicate the covariance between two inputs $x_i$ and $x_j$. Meanwhile, the NTK of a PIRBN suffices the following properties:

**Theorem 3.1.3.** *As $d \to \infty$, the NTK of PIRPN at the initialisation converges to a deterministic kernel $\mathbf{K}'$*

$$\lim_{d \to \infty} \mathbf{K}(0) \to \mathbf{K}'. \tag{22}$$

*Proof.* The proof is given in Appendix B.2.

**Theorem 3.1.4.** *As $d \to \infty$, assume that there exists a constant $\mathcal{C}$ such that*

$$\sup_{t \in [0,T]} \|\mathbf{a}(t)\|_\infty \leq \mathcal{C},$$

$$\sup_{t \in [0,T]} \|\mathbf{b}(t)\|_\infty \leq \mathcal{C},$$

$$\int_0^T \left| \sum_i^{n_g} \left( \mathrm{G}[u(x_i^g, \boldsymbol{\theta}(\sigma))] - g(x_i^g) \right) \right| d\sigma \leq \mathcal{C},$$

$$\int_0^T \left| \sum_i^{n_b} \left( \mathrm{B}[u(x_i^b, \boldsymbol{\theta}(\sigma))] - b(x_i^b) \right) \right| d\sigma \leq \mathcal{C},$$

*during the training process, i.e. $t \in [0, T]$. Then, the NTK of PIRBN through the gradient descendent algorithm suffices*

$$\lim_{d \to \infty} \sup_{t \in [0,T]} \|\mathbf{K}(t) - \mathbf{K}(0)\|_2 = 0. \tag{23}$$

*Proof.* The proof is given in Appendix B.3.

**Remark 3.1.5.** Based on the above Theorems, as $d \to \infty$, one can conclude that the NTK of a PIRBN remains unchanged during the training of the PIRBN via gradient descendent algorithms. In fact, when the width of a PIRBN is enough large, the NTK of the PIRBN also approximately remains static during training. Hence, training a PIRBN with enough large width through gradient descendent algorithms can also be regarded as solving a linear PDE problem, as stated in Eq. (10). Meanwhile, the NTK-based adaptive training scheme proposed by Wang et al. [50] is applicable for PIRBN with gradient descendent training algorithms.

By using the PIRBN with the NTK-based adaptive training scheme, we recall the problem Eq. (15) with $\mu = 8$. A PIRBN is set up with also 61 neurons. The centres of PIRBN neurons are uniformly distributed within [-99.9, 101.1] and initial $b = 10$. **Fig. 6** shows the results from PIRBN after $2 \times 10^4$ iterations via Adam optimiser (learning rate 0.001). The prediction obtained from the PIRBN aligns well with the analytical solution. Besides, the maximum point-wise absolute error is $7.31 \times 10^{-3}$. The loss terms converge after $1.5 \times 10^4$ iterations. Additionally, it is clear to find that the normalised NTK of the PIRBN remains as a diagonal matrix at the initial stage (0$^{th}$ iteration), 2000$^{th}$ iteration and 20000$^{th}$ iteration, suggesting that the PIRBN exhibits a well-local approximation property throughout the whole training process.

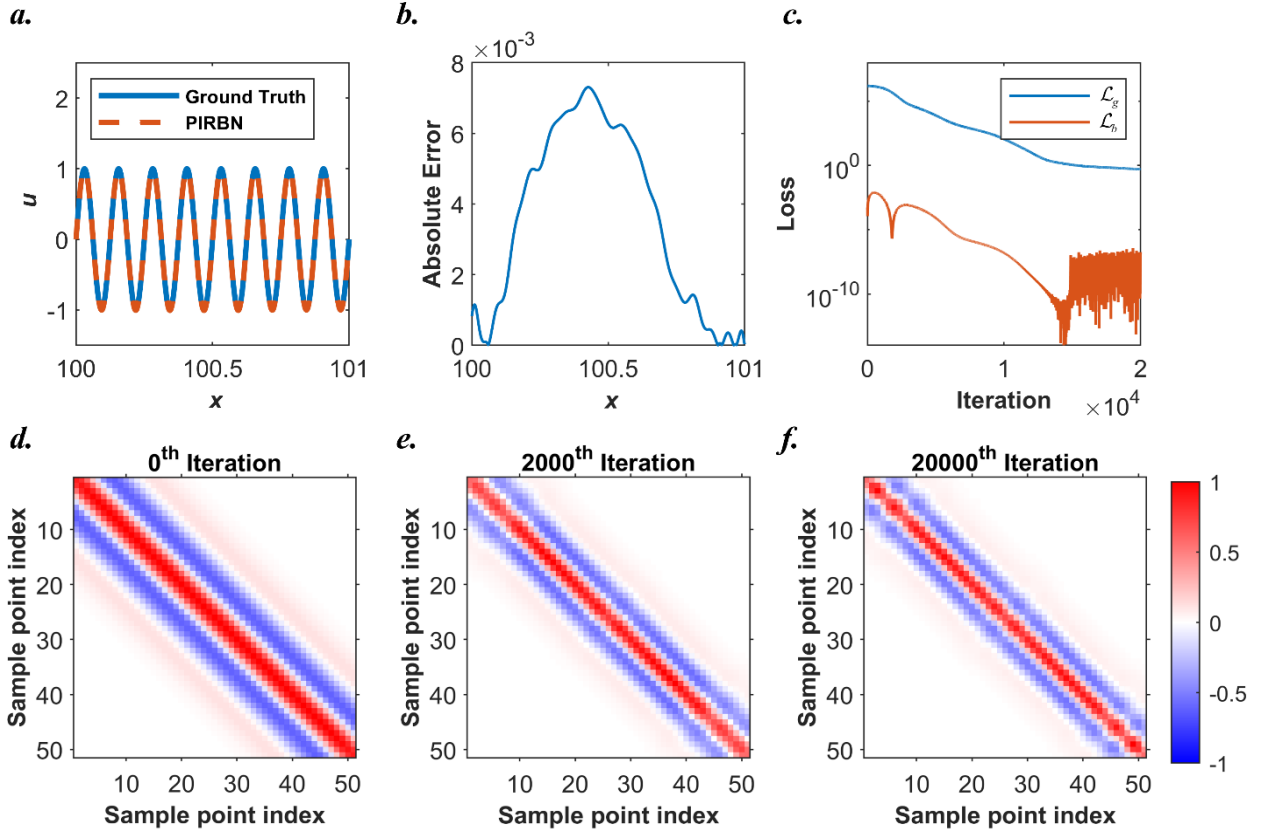

**Fig. 6.** Results from a PIRBN (single hidden layer with 61 neurons per layer) for solving Eq. (15) when $\mu = 8$. (a) Comparisons between the PIRBN predictions and the ground truth. (b) Point-wise absolute error plot. (c) Loss history of the PIRBN during the training process. (d) The normalised $\mathbf{K}_g$ at 0$^{th}$ iteration; (e) The normalised $\mathbf{K}_g$ at 2000$^{th}$ iteration; (f) The normalised $\mathbf{K}_g$ at 20000$^{th}$ iteration.

### 3.2. Parameter initialisation

In this part, we discuss the parameter initialisation for a PIRBN. For the sack of simplicity, the 1D PDE problem state in Eq.(15) is used as an example and the Gaussian function is selected for the PIRBN. As mentioned in the previous section, to analyse the training process of PIRBN via the NTK theory, weights in the second layer, $a_i$, require to be initialised as i.i.d. $\mathcal{N}(0,1)$ variables. Therefore, we only discuss the initialisation schemes for PIRBN with respect to $b_i$ and $c_i$.

We first start with the parameter $b$. As aforementioned, $b$ controls the width of the Gaussian function. Given that the Gaussian function has the following properties

$$\frac{\int_{-3/|b|}^{3/|b|} \vartheta(x)\mathrm{d}x}{\int_{-\infty}^{+\infty} \vartheta(x)\mathrm{d}x} \approx 99.8\%. \tag{24}$$

Therefore, we define the impact area of a Gaussian function neuron as $[c - 3/|b|, c + 3/|b|]$. Consequently, for 1D uniformly distributed sample points, the number of sample points within a Gaussian function, $\delta$, can be calculated as

$$\delta = 2 \left\lfloor \frac{3}{|b| \cdot dx} \right\rfloor + 1, \tag{25}$$

where $dx$ denotes the sample points spacing. $\delta$ can greatly influence the convergence rate of PIRBN's training process and the accuracy of PIRBN's prediction. Examples of PIRBNs initialised with different $b$ are presented in **Fig. 7**. As observed, with larger initial $b$, the NTK of the PIRBN exhibits poor diagonal property, suggesting that sample points inside the computational domain are still highly coupled with each other during the training process. On contrary, with decreasing initial $b$, the NTK of the PIRBN exhibits a better diagonal property. However, too large initial $b$ can induce insufficient sample points within the impact area of a Gaussian function neuron. In this case, the PIRBN will fail to govern the area between sample points, resulting in low accuracy when solving PDEs. This can be further demonstrated by Fig. 8, which presents histories of loss and mean absolute error (MAE) of PIRBNs with different initialised $b$ for solving Eq. (15). With increasing $b$, the loss and MAE of PIRBNs converge faster. However, despite the fastest convergence rate of loss shown in Fig. 8(a) when $b$ is initialised as 100, the MAE of the PIRBN shown in Fig. 8(b) does not decrease and remains at a high level during the training process. Meanwhile, the loss of the PIRBN with initial $b = 100$ jumps up after $1 \times 10^4$ iterations. This is because only insufficient sample points are available in the impact domain when $b$ is initialised as 100 and the Gaussian function neurons are poorly trained by insufficient support, comparisons between b = 100 and b = 25 are shown in Fig. 9. Considering both accuracy and efficiency, $b = 25$ is suggested for this problem.

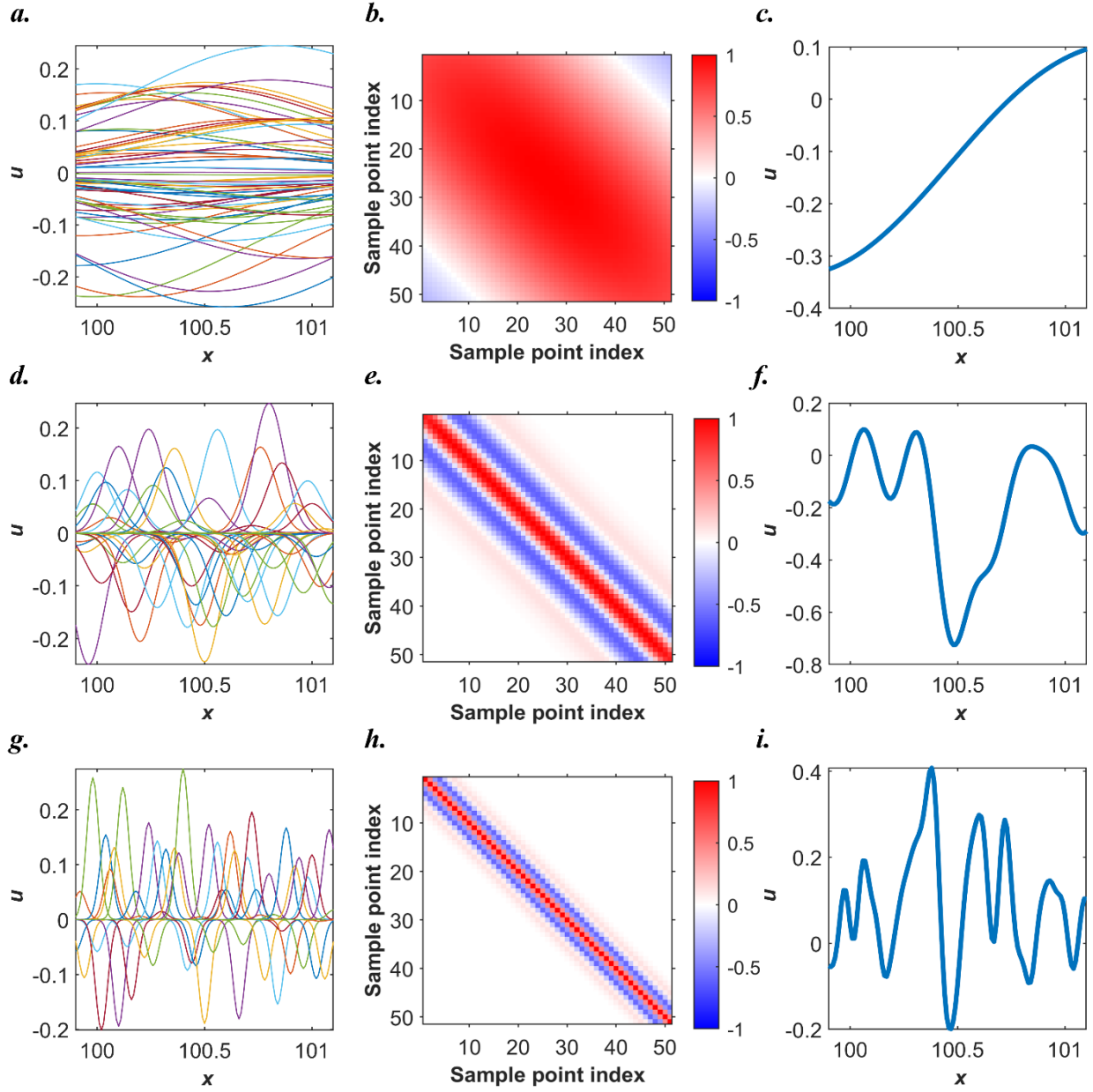

**Fig. 7.** Initialise a PIRBN with different $b$. Centres of RBF neurons are uniformly distributed within [99.9, 101.1]. (a) All Gaussian function neurons plot when initial $b = 1$. (b) The normalised $\mathbf{K}_g$ at the initial stage when initial $b = 1$. (c) The prediction of the PIRBN at the initial stage when initial $b = 1$. (d) All Gaussian function neurons plot when initial $b = 10$. (e) The normalised $\mathbf{K}_g$ at the initial stage when initial $b = 10$. (f) The prediction of the PIRBN at the initial stage when initial $b = 10$. (g) All Gaussian function neurons plot when initial $b = 25$. (h) The normalised $\mathbf{K}_g$ at the initial stage when initial $b = 25$. (i) The prediction of the PIRBN at the initial stage when initial $b = 25$.

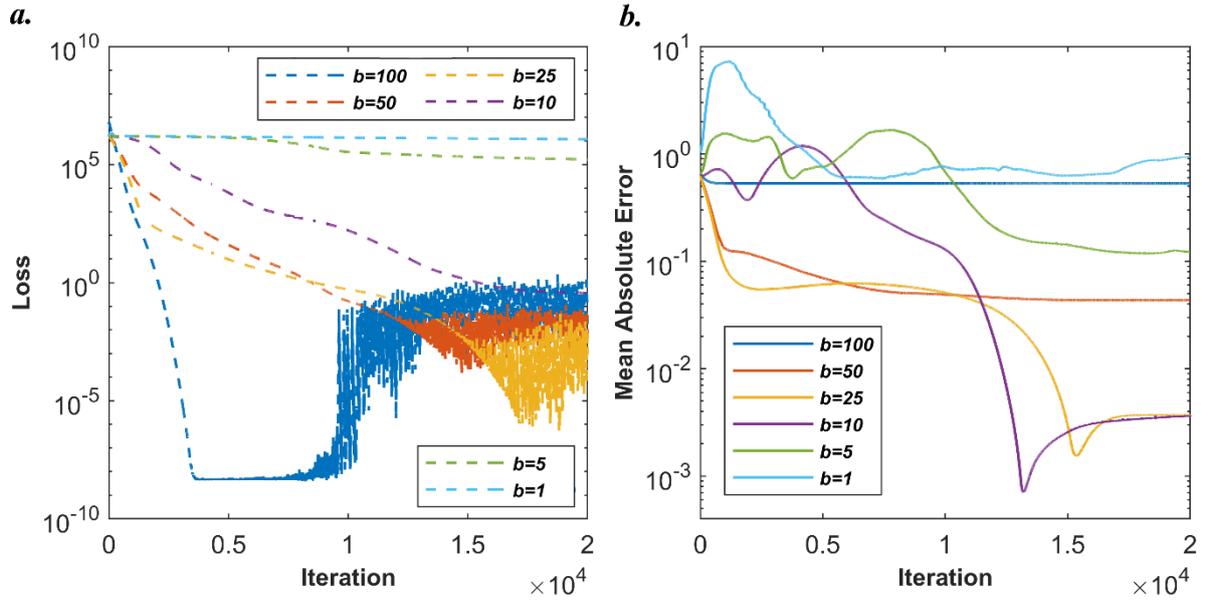

**Fig. 8.** (a) Loss history of PIRBNs initialised with different initial *b*. (b) Mean absolute error history of PIRBNs with different initial *b*.

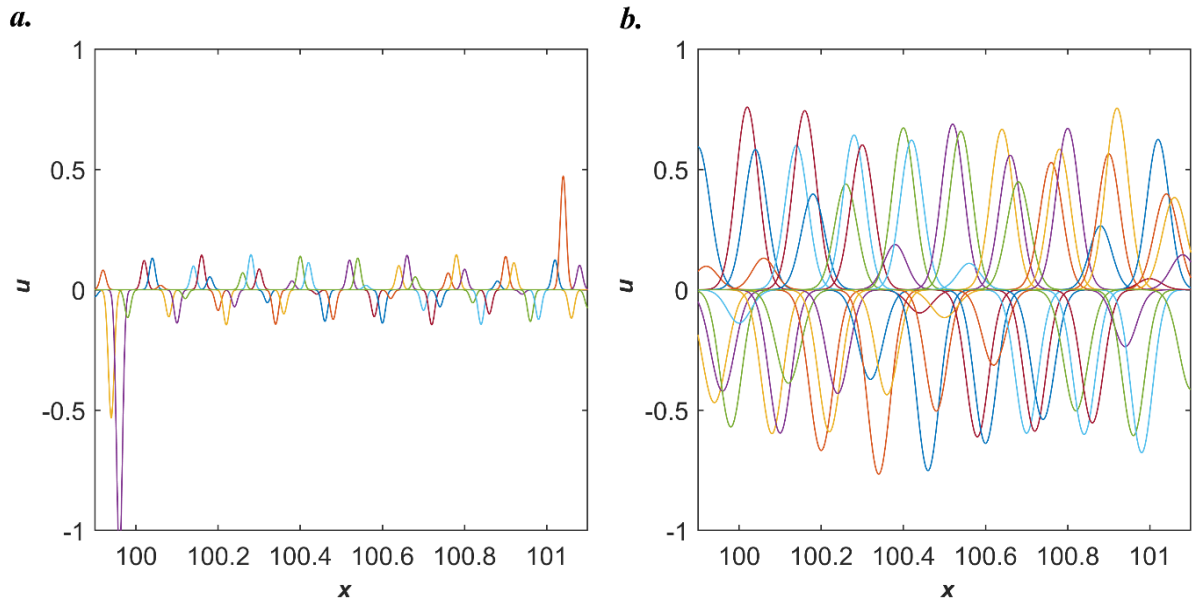

**Fig. 9.** RBF neurons plot of PIRBNs after training. (a) Initial $b = 100$; (b) Initial $b = 25$.

Next, we discuss the initialisation for the centre of each RBF neuron. As aforementioned, *c* controls the location of the centre of an RBF neuron. Given that the computational domain for most PINN applications is bounded, most of the centres of RBF neurons in a PIRBN should be initialised inside the computational domain. Specifically, extra Gaussian function neurons whose centres are around the outer edge of the boundaries are required, as shown in Fig. 10(a). By adding those outer edge RBF neurons, PIRBN has an increasing capability of approximating the boundary conditions. Compared to PIRBN without the outer edge neurons shown in Fig. 10(b), the $\mathcal{L}_b$ of the PIRBN with outer edge neurons can converge faster and reach a lower level, as shown in Fig. 10(c). Based on our numerical experimental experiences, 4~10 outer edge neurons are suggested.

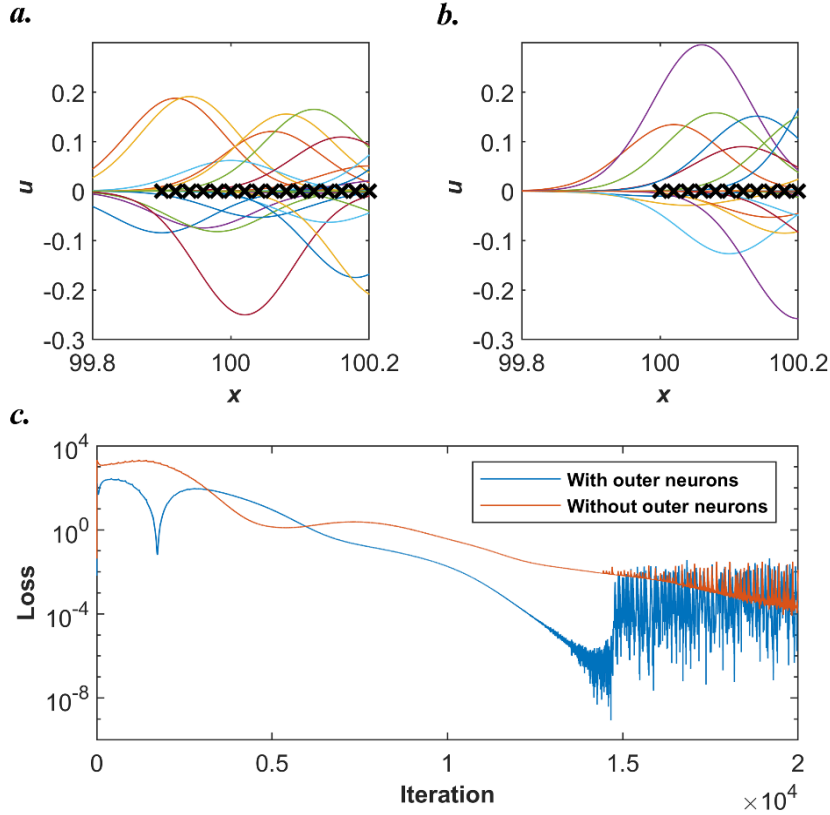

**Fig. 10.** (a) Initialise PIRBN with extra Gaussian function neurons at the outer edge of the computational domain. (b) Initialise PIRBN without extra Gaussian function neurons at the outer edge of the computational domain. The black crosses denote the centres of the Gaussian function neurons. (c) Loss history of $\mathcal{L}_b$ during the training process.

Till now, only PIRBNs with uniformly distributed Gaussian function neurons are considered within the computational domain. We further test the performance of PIRBNs when using randomly distributed Gaussian function neurons. The initial $b$ for all cases is set to be 10 so that $\delta$ for all cases is no less than 15. **Fig. 11** shows the results from PIRBNs by using uniformly and randomly distributed Gaussian function neurons. Compared with the PIRBN with uniformly distributed centres, the PIRBN with randomly distributed centres exhibits poor performance. The maximum point-wise absolute error produced by PIRBN with randomly distributed centres is 0.011, while the maximum point-wise absolute error produced by PIRBN with uniformly distributed centres is 0.007. **Table 1** lists the mean values of $\mathcal{L}_g$, $\mathcal{L}_b$ and MAE by using PIRBNs with randomly and uniformly distributed centres. As observed, both losses and MAE of the PIRBN with uniformly distributed centres reach significantly lower levels than the PIRBN with randomly distributed centres, suggesting that the randomly distributed centres significantly influence the performance of the PIRBN. In this manner, uniformly distributed Gaussian function neurons are highly suggested if applicable.

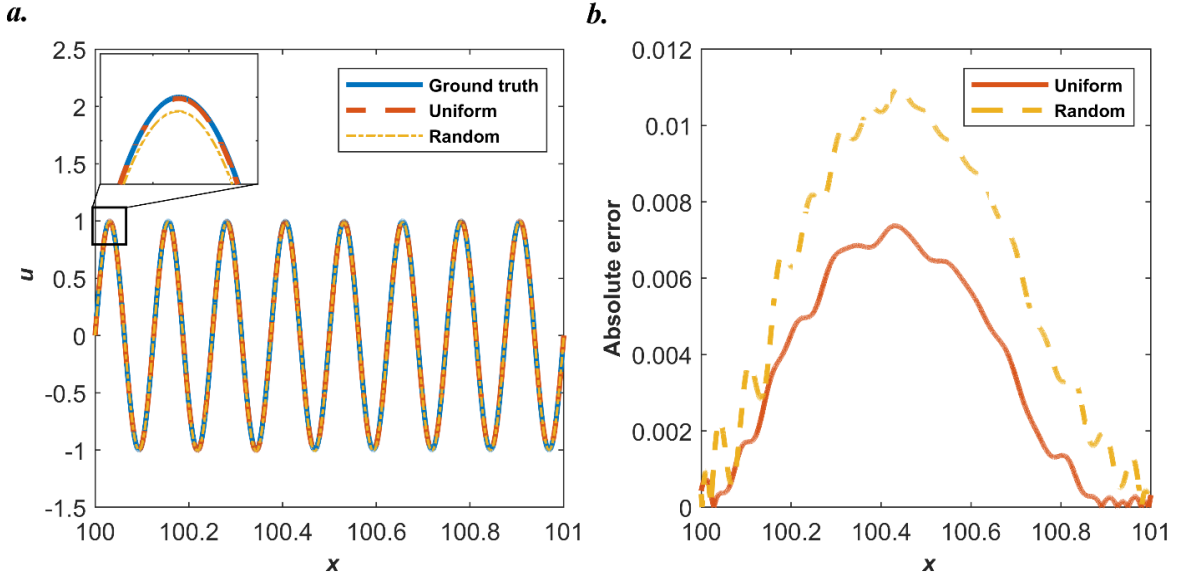

**Fig. 11.** Results from PIRBNs by using uniformly and randomly distributed Gaussian function neurons. (a) Comparisons between the analytical solution and predictions from PIRBNs by using uniformly and randomly distributed Gaussian function neurons. (b) The point-wise absolute error of predictions from PIRBNs.

**Table 1.** Comparisons of $\mathcal{L}_g$, $\mathcal{L}_b$ and mean absolute error between PIRBNs with randomly and uniformly distributed centres. Initial $b$ is set to be 10. The values are obtained from 20 times runs.

|  | $\mathcal{L}_g$ | $\mathcal{L}_b$ | Mean absolute error (MAE) |
|---|---|---|---|
| Random | $23.99 \pm 32.31$ | $2.21 \pm 3.96 \times 10^{-3}$ | $7.55 \pm 7.10 \times 10^{-3}$ |
| Uniform | $0.46 \pm 0.14$ | $4.51 \pm 8.24 \times 10^{-9}$ | $3.62 \pm 0.12 \times 10^{-3}$ |

*3.3. Number of sample points*

To test the performance of PIRBN with different sample points inside the computational domain, 5 different numbers of uniformly distributed sample points are used to solve Eq.(15). For all the cases, 601 Gaussian function neurons are uniformly distributed within [99.9, 101.1]. Different initial $b$ values are set for different cases so that $\delta \approx 21$ for all cases.

**Fig. 12** shows the results from PIRBN with different numbers of uniformly distributed sample points. As shown in **Fig. 12**(a), the predictions from PIRBNs with different numbers of sample points agree well with the analytical solution, suggesting that all the cases can effectively solve the problem. **Fig. 12**(b) shows the MAE history of PIRBNs with different numbers of sample points throughout the training processes. As observed, the accuracy of PIRBN can be increased with the increasing number of sample points. However, it is also clear to find, with the same learning rate, using more sample points requires more iterations for convergence. Considering that more sample points can make the PIRBN training more sensitive, too large a learning rate may be difficult for convergence. We further test the influence of the learning rate for PIRBN training, and the convergence plot is presented in Fig. 12(c). The learning rates of the six plots are $10^{-2}$, $10^{-3}$, $10^{-4}$, $10^{-5}$, $10^{-6}$ and $10^{-10}$. As observed, when using the learning rate $10^{-2}$, the training of the PIRBN is unstable and fails to converge. When the learning rate is smaller than $10^{-2}$, the training of a PIRBN can converge earlier with the decreasing learning rate till $10^{-6}$. Nevertheless, applying too small a learning rate can lead to slower convergence, as the convergence plot for $10^{-10}$. Hence, the selection of a proper learning rate can enhance PIRBNs' training efficiency.

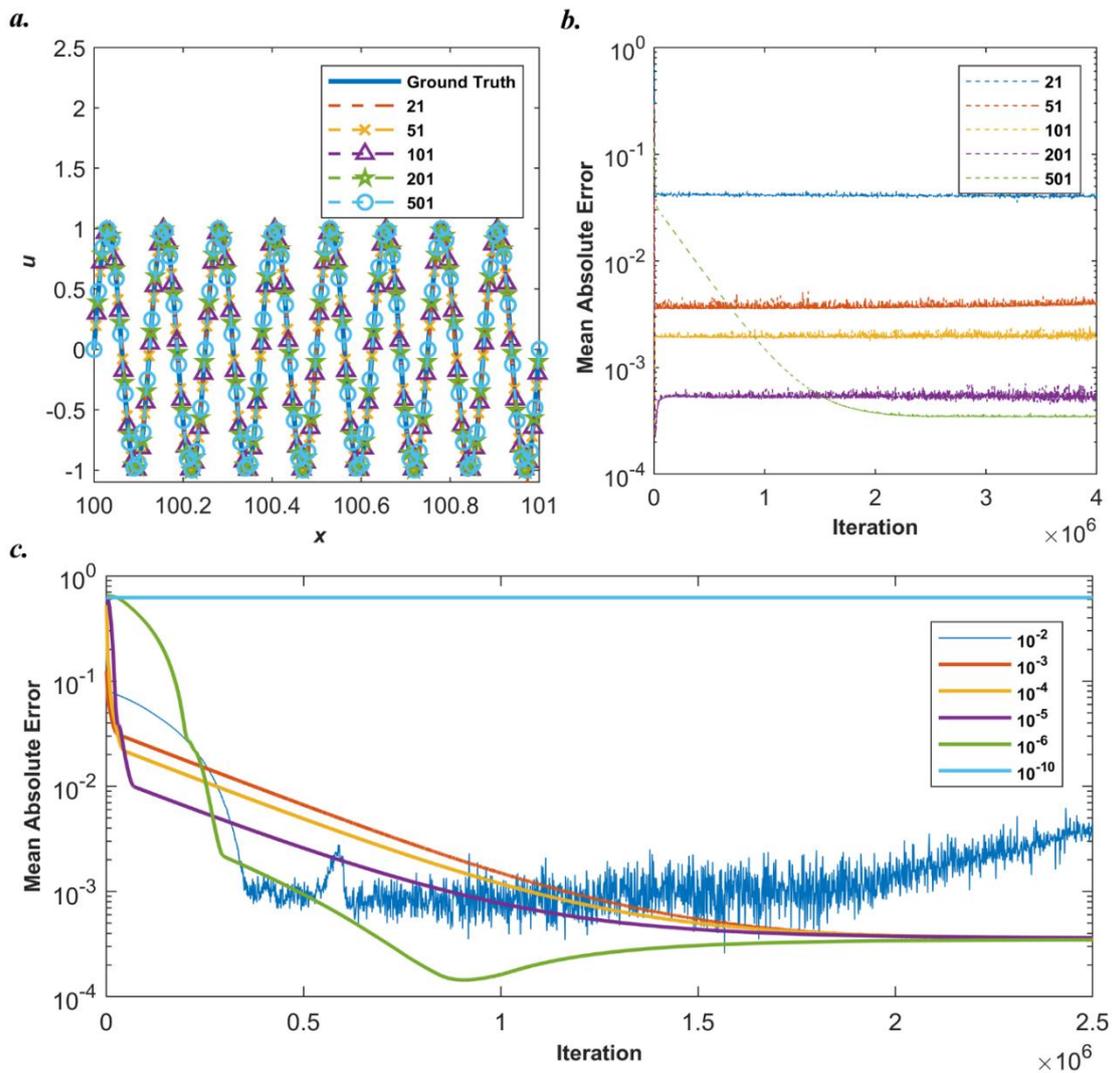

**Fig. 12.** Results from PIRBNs by using different numbers of uniformly distributed sample points. (a) Comparisons between the analytical solution and predictions from PIRBNs with different numbers of uniformly distributed sample points. (b) MAE history of PIRBNs with different numbers of sample points throughout training processes. (c) Convergence plots for PIRBN training with learning rates as $10^{-2}$, $10^{-3}$, $10^{-4}$, $10^{-5}$, $10^{-6}$ and $10^{-10}$.

*3.4. Selection of RBF*

As previously mentioned, there are various types of RBF. However, not all types of RBF are effective for constructing a PIRBN. Considering the prerequisites in the theorems, the RBF used in PIRBN should satisfy the following requirements:

i) The selected RBF should be high-order differentiable (infinitely smooth).
ii) The selected RBF and its derivatives should be bounded.
iii) The selected RBF should have a local impact.

Herein, we choose four typical RBFs and test performances of PIRBN with different RBFs, including the Gaussian function, the inverse quadratic function, the inverse multiquadric function and the thin plate spline, as shown in **Fig. 13**. Apart from the Gaussian function, expressions of the rest RBFs are provided as

$$\text{Inverse quadratic function}: \quad R_{iq}(\|\mathbf{x}-\mathbf{c}\|) = \frac{1}{1+b^2\|\mathbf{x}-\mathbf{c}\|^2}, \quad (26)$$

$$\text{Inverse multiquadric function}: \quad R_{imq}(\|\mathbf{x}-\mathbf{c}\|) = \frac{1}{\sqrt{1+b^2\|\mathbf{x}-\mathbf{c}\|^2}}, \quad (27)$$

$$\text{Thin plate spline}: \quad R_{tp}(\|\mathbf{x}-\mathbf{c}\|) = b^2\|\mathbf{x}-\mathbf{c}\|^2 \ln(\sqrt{b^2\|\mathbf{x}-\mathbf{c}\|^2}+1). \quad (28)$$

It is clear to find that only the thin plate spline does not fully satisfy the requirements, for the thin plate spline and its derivatives are neither bounded nor have local impact.

Now, consider a PDE and its boundary conditions

$$\begin{aligned}\frac{d^2}{dx^2}u(x) &= f(x), \quad \text{for } x \in [20,22], \\ u(20) &= u(22) = 0,\end{aligned} \quad (29)$$

where *f(x)* is a tailored function so that the solution of this PDE can be written as

$$u(x) = \left(\frac{22-x}{2}\right)^2 \sin(2\pi x) + \left(\frac{x-20}{2}\right)^2 \sin(16\pi x). \quad (30)$$

Again, this problem is challenging for PINN since the computational domain is not normalised. Besides, the analytical solution contains both low-frequency and high-frequency features; that is, the solution exhibits a low-frequency feature when *x* is closer to 20, while the solution exhibits a high-frequency feature when *x* is approaching 22. **Fig. 14** shows the results obtained by using PINN (one hidden layer and 121 neurons in the hidden layer). As observed, PINN fails to solve this problem by using the Adam optimiser (learning rate 0.001) after $2\times10^4$ iterations. The loss terms from the PDE can be hardly decreased during the training process, as shown in **Fig. 14**(c).

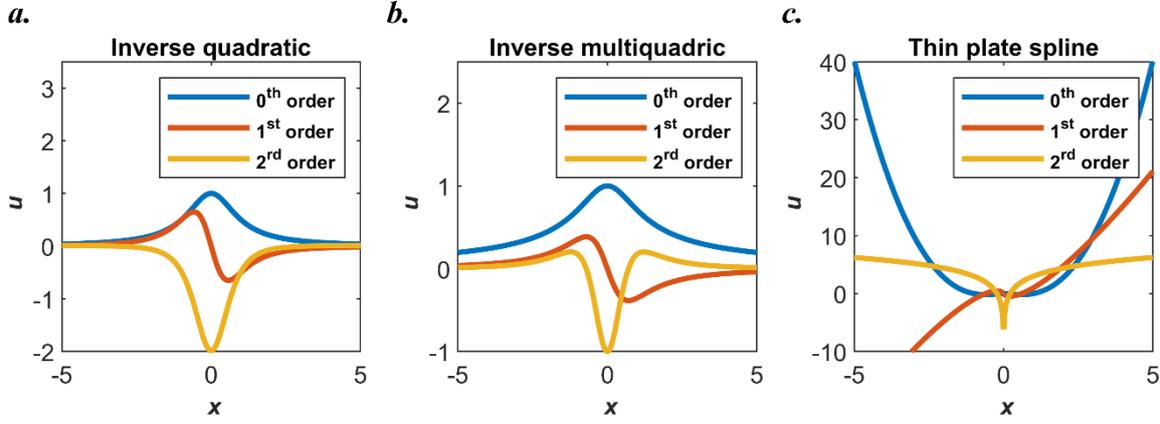

**Fig. 13.** Illustrations of some typical RBFs. (a) The inverse quadratic function; (b) The inverse multiquadric function; (c) The thin plate spline.

**Fig. 15**, **Fig. 16** and **Fig. 17** show the results obtained by using PIRBNs (121 uniformly distributed neurons) with the Gaussian function, the inverse quadratic function and the inverse multiquadric function, respectively. All these PIRBNs can effectively solve the problem with good accuracy via the Adam optimiser after $2\times10^4$ iterations. Besides, NTKs for all these PIRBNs retain as diagonal matrixes throughout the training processes, as shown in **Fig. 15**(d)-(f), **Fig. 16**(d)-(f) and **Fig. 17**(d)-(f). Among them, the PIRNB with the Gaussian function performs the best, while the largest point-wise absolute error is $1.15 \times 10^{-3}$.

Fig. 18 shows the results obtained by using a PIRBN with the thin plate spline. As observed from Fig. 18(a) and (b), the PIRBN fails to solve the problem, suggesting that not all RBFs can be effective for constructing a PIRBN. In terms of the loss history of the PIRBN with the thin plate spline plotted in Fig. 18(c), all loss terms remain at a high level during the training process, suggesting that the PIRBN with the thin plate spline cannot be well-trained. This is further proved by the MAE plot which remains nearly unchanged throughout the whole training process, as also shown in Fig. 18(c).

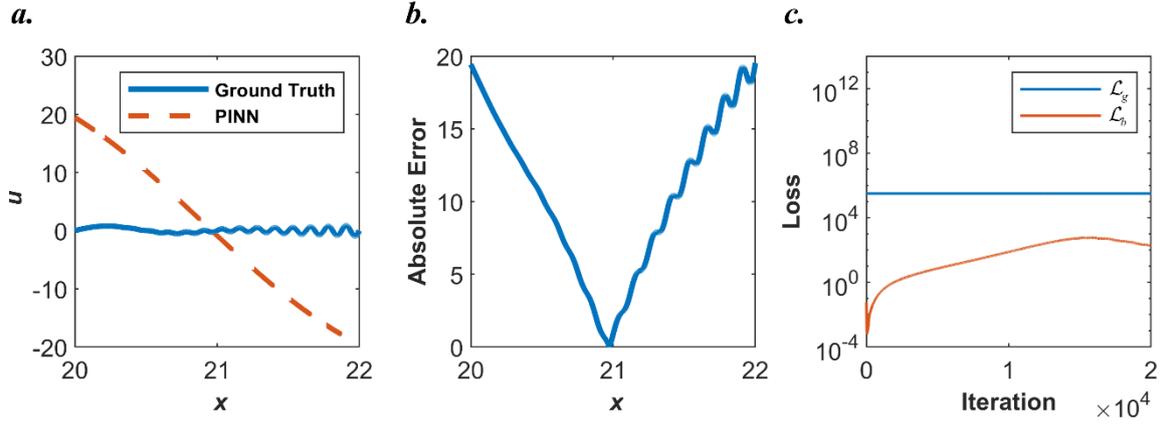

**Fig. 14.** Results from a PINN (single hidden layer with 61 neurons per layer) for solving Eq. (29). (a) Comparisons between the predictions from the PINN and the analytical solution. (b) Point-wise absolute error from the PINN. (c) Loss history of the PINN during the training process.

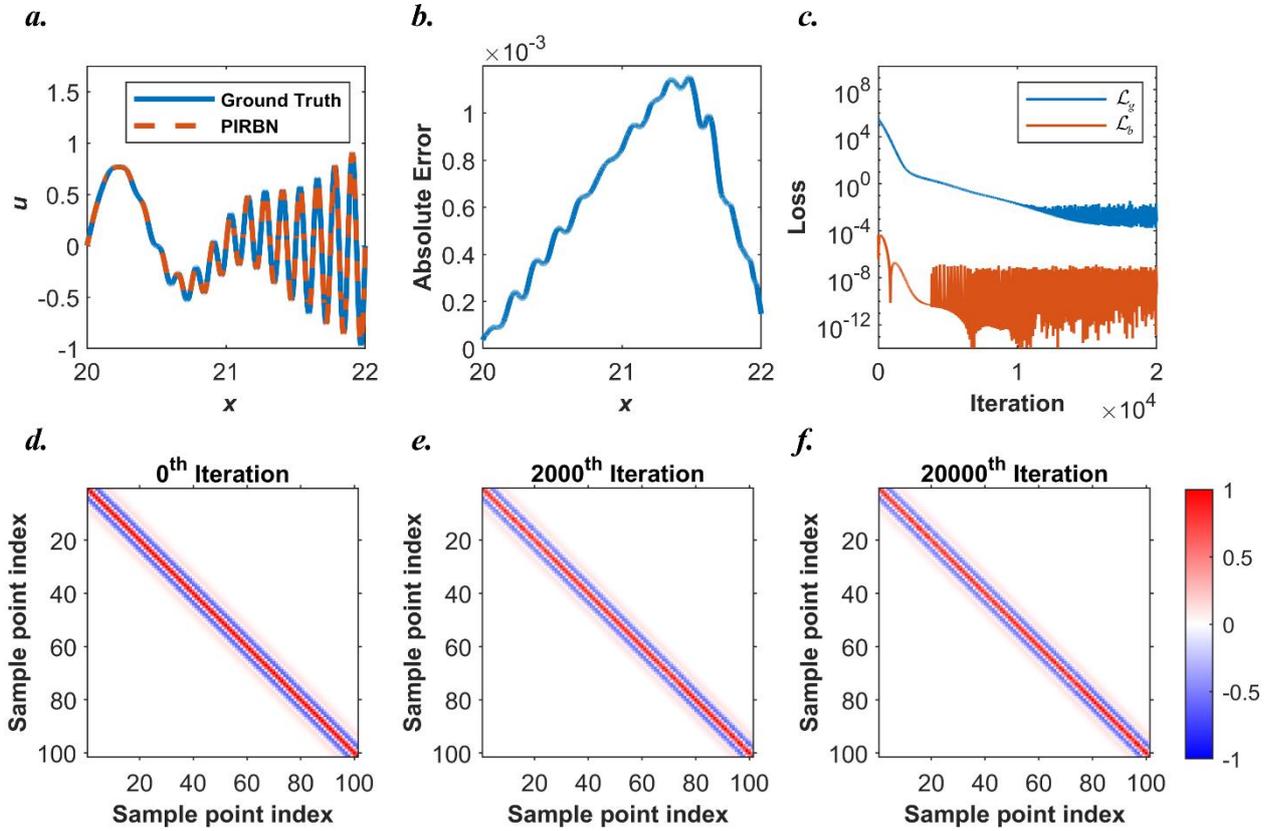

**Fig. 15.** Results from a PIRBN (single hidden layer with 101 neurons per layer) with the Gaussian function for solving Eq. (29). (a) Comparisons between the predictions from the PIRBN and the analytical solution. (b) Point-wise absolute error from the PIRBN. (c) Loss history of the PIRBN during the training process. (d) The normalised $\mathbf{K}_g$ at the $0^{th}$ iteration. (e) The normalised $\mathbf{K}_g$ at the $2000^{th}$ iteration; (f) The normalised $\mathbf{K}_g$ at the $20000^{th}$ iteration.

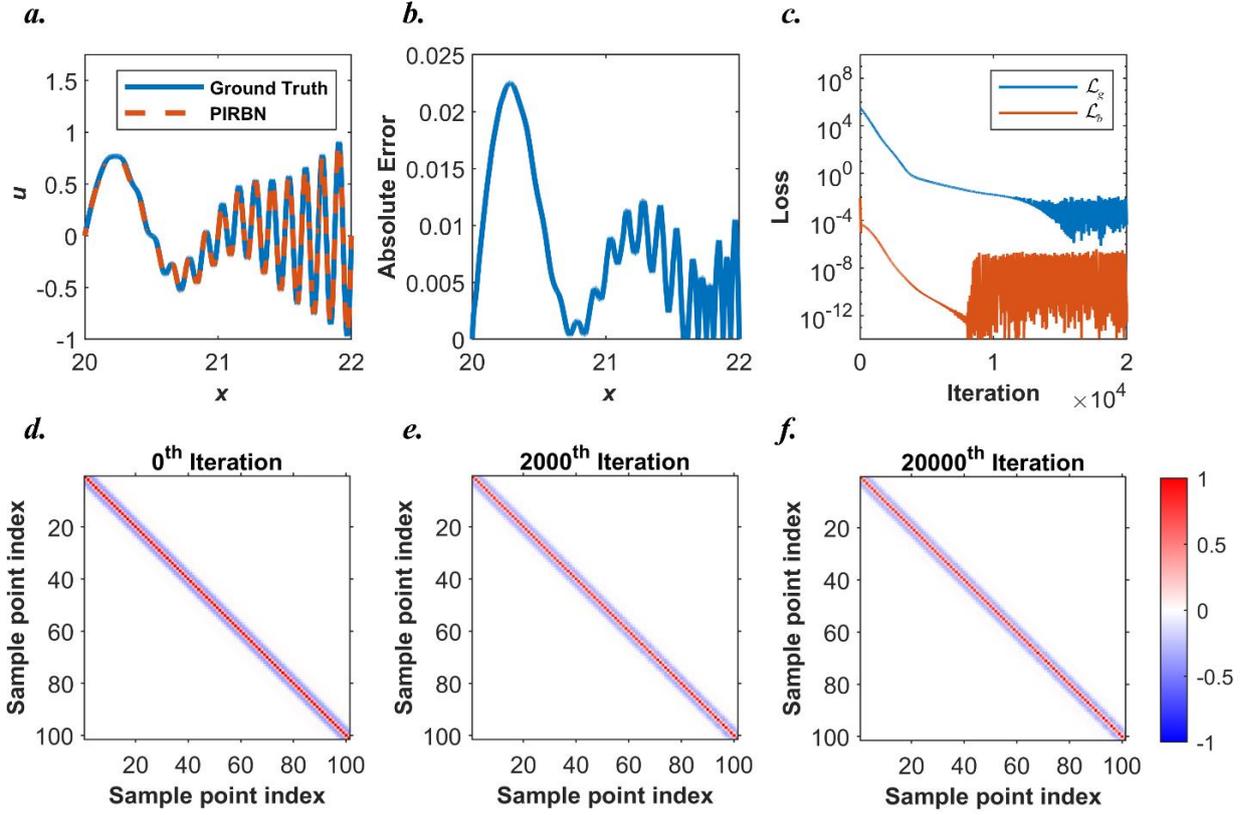

**Fig. 16.** Results from a PIRBN (single hidden layer with 101 neurons per layer) with the inverse quadratic function for solving Eq. (29). (a) Comparisons between the predictions from the PIRBN and the analytical solution. (b) Point-wise absolute error from the PIRBN. (c) Loss history of the PIRBN during the training process. (d) The normalised $\mathbf{K}_g$ at the 0$^{th}$ iteration. (e) The normalised $\mathbf{K}_g$ at the 2000$^{th}$ iteration; (f) The normalised $\mathbf{K}_g$ at the 20000$^{th}$ iteration.

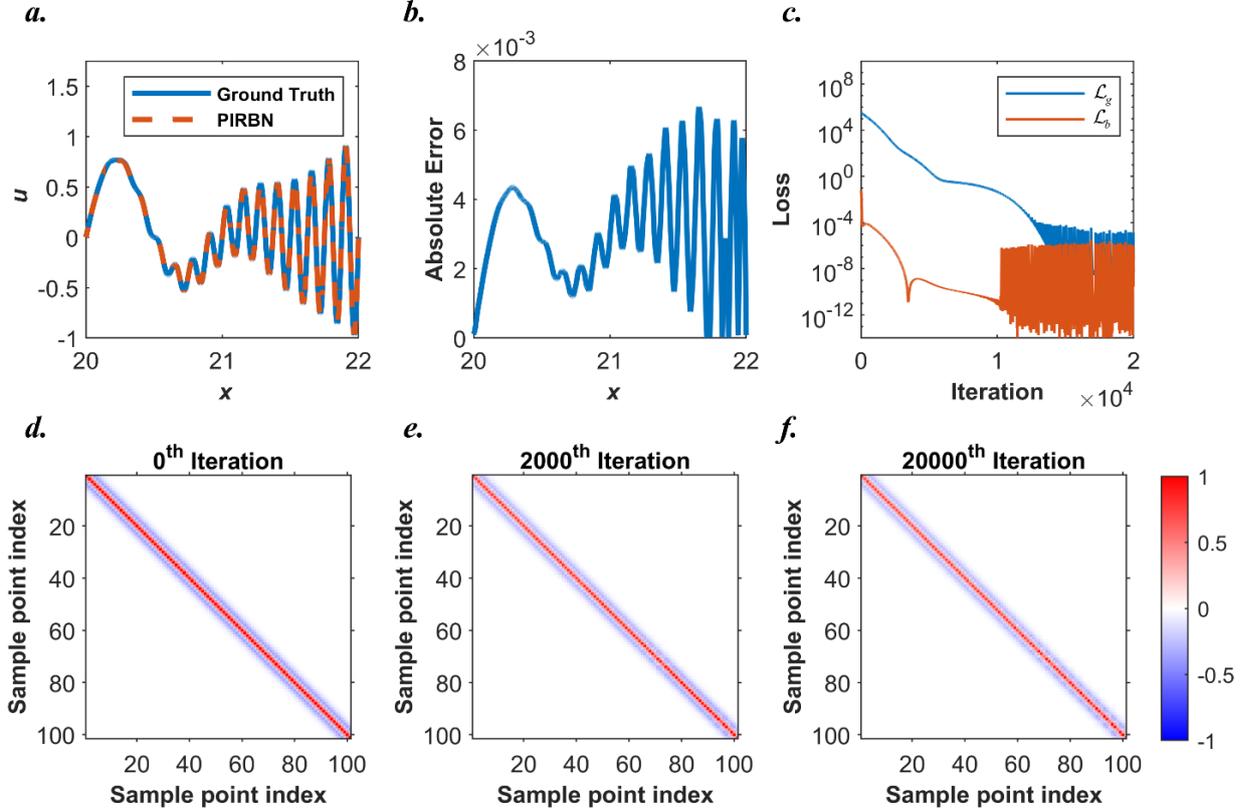

**Fig. 17.** Results from a PIRBN (single hidden layer with 101 neurons per layer) with the inverse multiquadric function for solving Eq. (29). (a) Comparisons between the predictions from the PIRBN and the analytical solution. (b) Point-wise absolute error from the PIRBN. (c) Loss history of the PIRBN during the training process. (d) The normalised $\mathbf{K}_g$ at the 0$^{th}$ iteration. (e) The normalised $\mathbf{K}_g$ at the 2000$^{th}$ iteration; (f) The normalised $\mathbf{K}_g$ at the 20000$^{th}$ iteration.

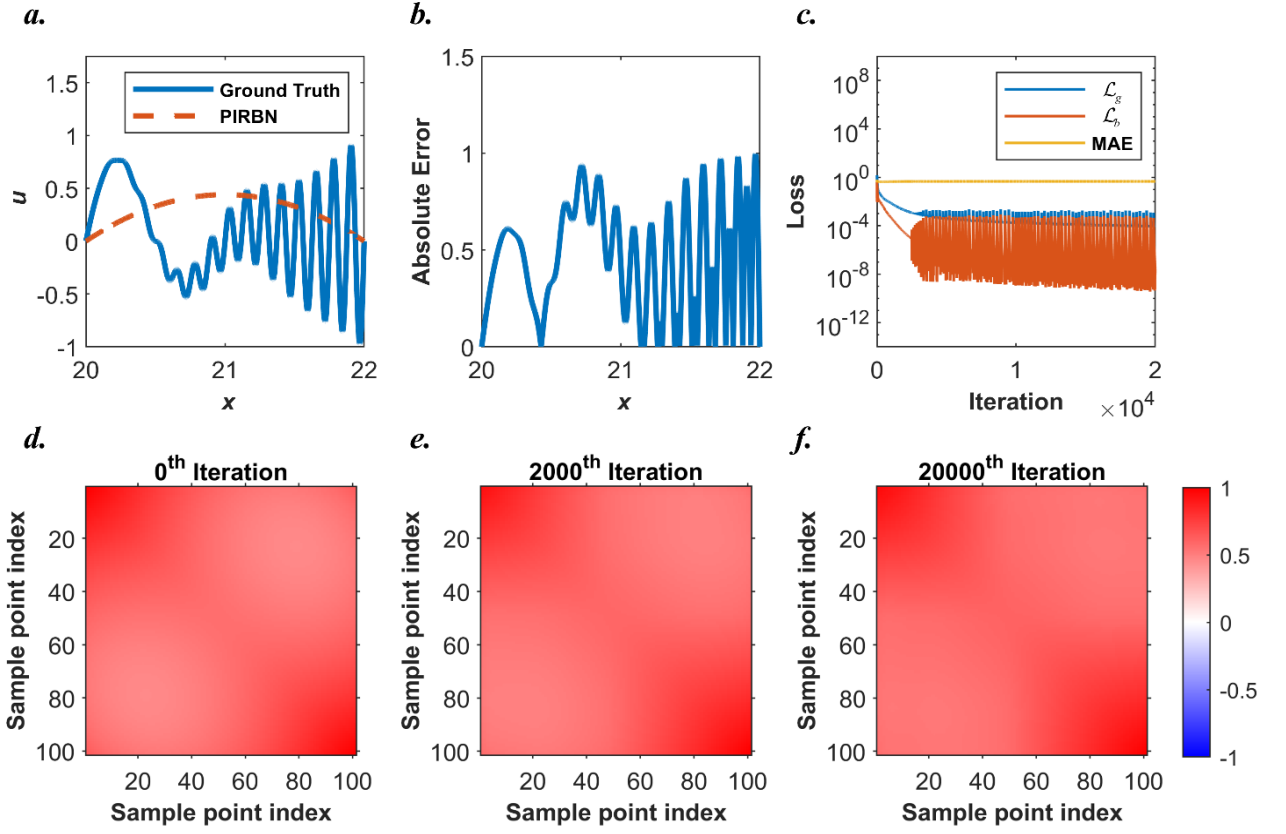

**Fig. 18.** Results from a PIRBN (single hidden layer with 101 neurons per layer) with the thin plate spline for solving Eq. (29). (a) Comparisons between the predictions from the PIRBN and the analytical solution. (b) Point-wise absolute error from the PIRBN. (c) Loss history of the PIRBN during the training process. (d) The normalised $\mathbf{K}_g$ at the 0$^{th}$ iteration. (e) The normalised $\mathbf{K}_g$ at the 2000$^{th}$ iteration; (f) The normalised $\mathbf{K}_g$ at the 20000$^{th}$ iteration.

## 3. Numerical examples

In this section, several numerical examples are conducted to demonstrate the performance of PIRBN towards different PDEs. All those PDEs are prevailingly used in various applications. We note that all the numerical examples are trained on a Windows 10 system with Intel(R) Core(TM) i7-8700 CPU@3.20GHz. The neural networks are built based on the TensorFlow library. The Adam optimiser is selected as the training algorithm. The programs that can regenerate all numerical results can be found at https://github.com/JinshuaiBai/PIRBN.

*3.1. 1D nonlinear spring equation*

Consider a 1D nonlinear spring equation and its initial/boundary conditions as follows

$$\frac{d^2}{dx^2}u(x) + 4u(x) + \sin(u(x)) = f(x), \text{ for } x \in [0,100],$$
$$u(0) = 0 \tag{31}$$
$$\frac{d}{dx}u(0) = 0$$

where *f(x)* is a tailored function so that the analytical solution of this problem can be written as

$$u(x) = x\sin(x). \tag{32}$$

This problem is quite challenging for PINN due to the exitances of a long-range computational domain $x \in [0,100]$, large predict value and high-frequency feature. To address this problem via PINN, Dong and Li [43] decomposed the whole computational domain into pieces and applied multiple neural networks to approximate the solution for each subdomain. Here, we apply a PIRBN with 1021 neurons to solve this problem. The centres of neurons are uniformly distributed within [-1,101], and the initial *b* is set to 10.

Fig. 19 shows the results from PINN and PIRBN for solving the 1D nonlinear spring equation, respectively. As observed from Fig. 19 (a) and (b), the predictions from the PIRBN align well with the analytical solution, while the largest point-wise absolute error is $8 \times 10^{-3}$. On contrary, the prediction from PINN only aligns the analytical results at the area close to *x* = 0, as shown in Fig. 19 (d) and (e). Meanwhile, the loss terms from PIRBN keep decreasing during the training process, while the loss terms from PINN remain nearly unchanged during the training process, as shown in Fig. 19 (c) and (f). Therefore, it is evident that the PIRBN is more effective for solving PDEs with high-frequency features and long-range domains compared to the original PINN.

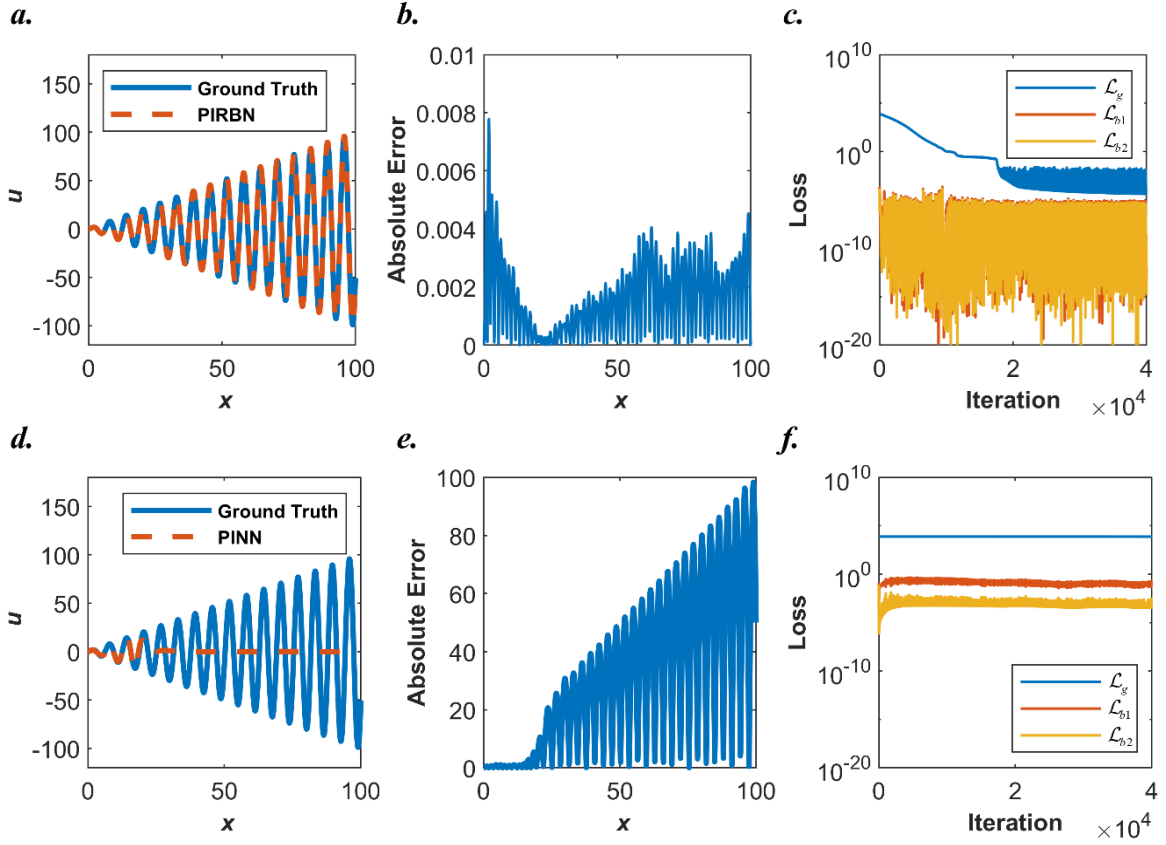

**Fig. 19.** Results for 1D nonlinear spring equation. (a) Comparison between the prediction from the PIRBN and the analytical solution. (b) Point-wise absolute error from the PIRBN. (c) Loss history of the PIRBN during the training process. (d) Comparison between the prediction from the PINN and the analytical solution. (e) Point-wise absolute error from the PINN. (f) Loss history of the PIRBN during the training process.

### 3.2. Wave equation

Next, we extend the use of PIRBN to 2D problems. Consider the following 2D wave equation and the corresponding boundary conditions as

$$\left(\frac{\partial^2}{\partial x^2} + 4\frac{\partial^2}{\partial y^2}\right)u(x,y) = 0, \quad \text{for } x \in [0,1], y \in [0,1],$$

$$u(x,0) = u(x,1) = \frac{\partial}{\partial x}u(0,y) = 0, \tag{33}$$

$$u(0,y) = \sin(\pi y) + \frac{1}{2}\sin(4\pi y).$$

The analytical solution is given as

$$u(x,y) = \cos(2\pi x)\sin(\pi y) + \frac{1}{2}\cos(8\pi x)\sin(4\pi y). \tag{34}$$

This problem is quite challenging due to the existence of high-frequency feature. Here, we apply a single-layer PIRBN with $61 \times 61$ neurons to solve this problem. The centres of neurons are uniformly distributed within computational domain $x \in [-0.1, 1.1]$, $y \in [-0.1, 1.1]$, where the spacing of neurons is 0.02. The initial $b$ is set to 20.

Fig. 20 shows the results from PIRBN for the wave equation. As observed, the predictions from PIRBN agree well with the analytical solution, while the maximum point-wise absolute error is $6 \times 10^{-4}$. The training process converges after approximately $8 \times 10^3$ iterations, as shown in **Fig. 20**(d). Besides, a PINN with a same size neural network is applied to solve this problem for comparison, and its results are shown in **Fig. 21**. It is clear to find that the PINN fails to predict the solution of the wave equation and its loss terms hardly change during the training process. It is also worth highlighting that, to address this problem via PINN, Wang et al.[50] applied an FNN with 5 hidden layers and 500 neurons per layer. Although a larger size FNN can produce good quality prediction, the maximum point-wise absolute error is still roughly $5 \times 10^{-3}$ [50], which is significantly larger than it was obtained by the PIRBN.

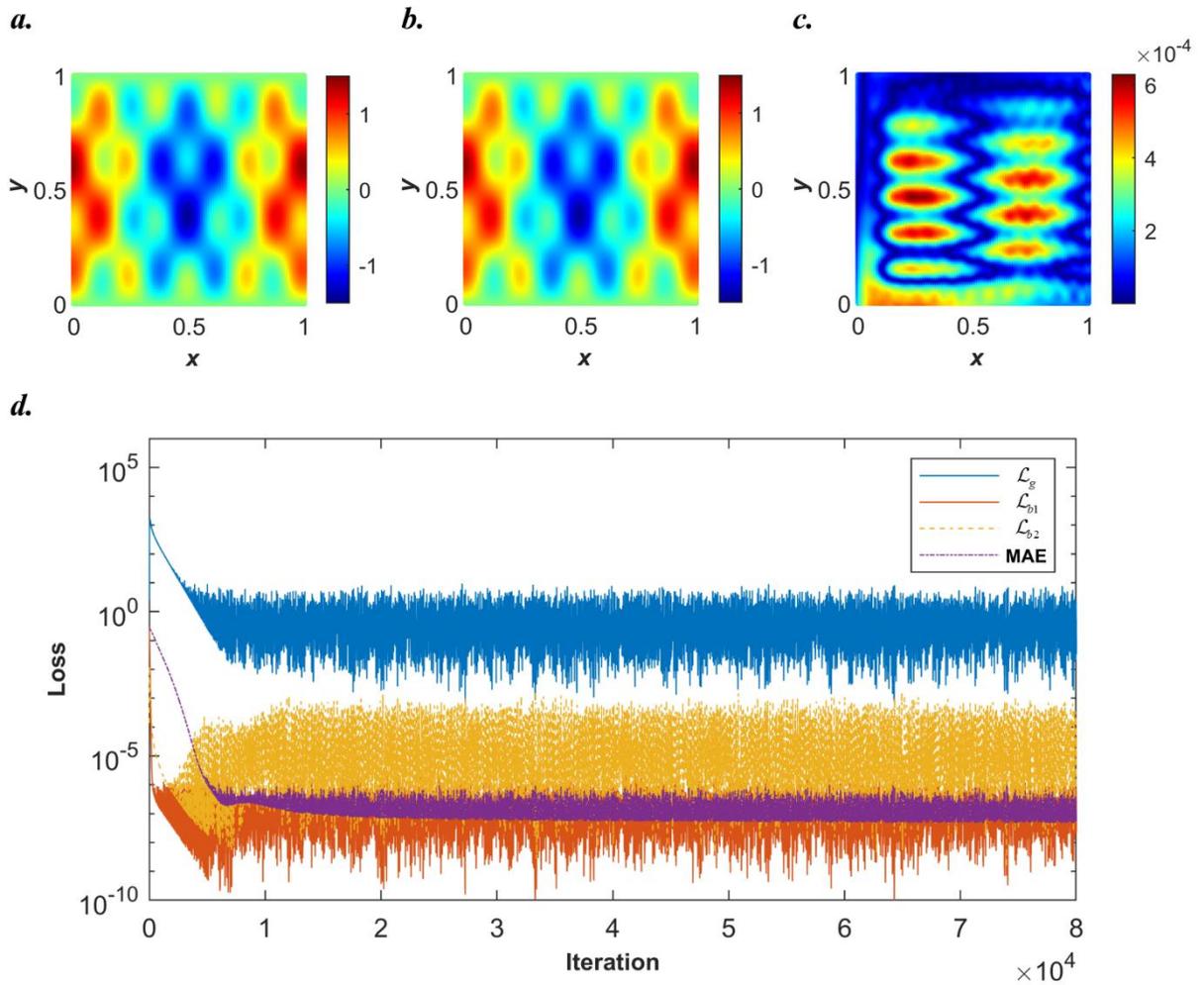

**Fig. 20.** PIRBN results for the 2D wave equation. (a) The prediction from the PIRBN. (b) The analytical solution. (c) Point-wise absolute error from the PIRBN. (d) Loss history and mean absolute error (MAE) of the PIRBN during the training process.

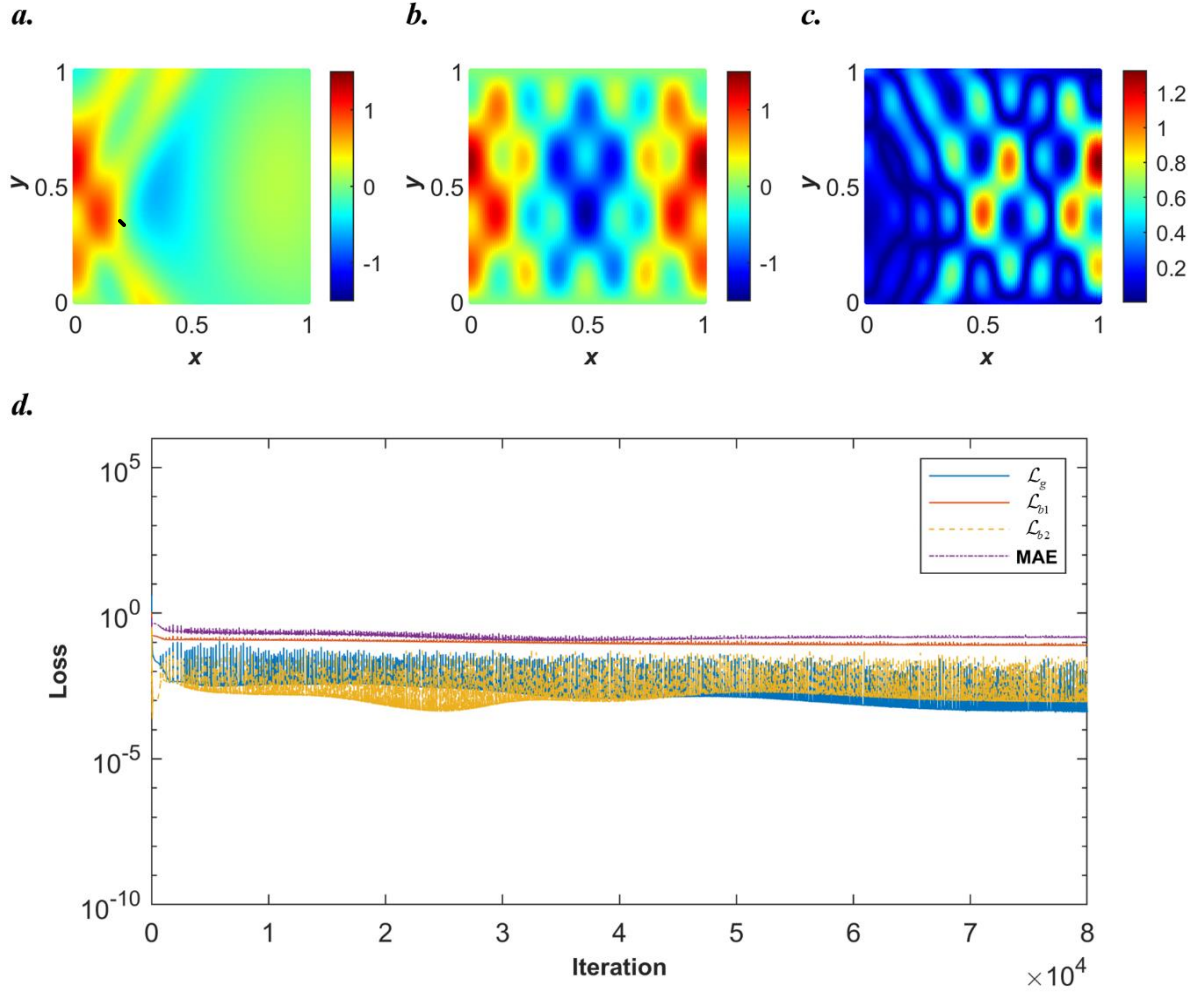

**Fig. 21.** PINN results for the 2D wave equation. (a) The prediction from the PINN. (b) The analytical solution. (c) Point-wise absolute error from the PINN. (d) Loss history and mean absolute error (MAE) of the PINN during the training process.

### 3.3. Diffusion equation

The diffusion equation is prevailingly seen in engineering, e.g., hydrodynamics [61] and heat transfer problems[62]. Consider a diffusion equation with its initial/boundary conditions as

$$\left(\frac{\partial}{\partial t} - 0.01\frac{\partial^2}{\partial x^2}\right)u(x,t) = g(x,t), \text{ for } x \in [5,10], t \in [5,10]$$
$$u(5,t) = b_1(t),$$
$$u(10,t) = b_2(t),$$
$$u(x,5) = b_3(x), \tag{35}$$

where $g(x, t)$, $b_1(t)$, $b_2(t)$ and $b_3(x)$ are tailored functions so that the solution $u(x, t)$ can be written as

$$u(x,t) = \left[2\cos\left(\pi x + \frac{\pi}{5}\right) + \frac{3}{2}\cos\left(2\pi x - \frac{3\pi}{5}\right)\right]\left[2\cos\left(\pi t + \frac{\pi}{5}\right) + \frac{3}{2}\cos\left(2\pi t - \frac{3\pi}{5}\right)\right]. \quad (36)$$

This problem is challenging for PINN due to the existence of the high-frequency feature and ill-posed computational domain. It is solved in [43] by domain decomposition, i.e., the whole computational domain is partitioned into subsections and local neural networks are used to respectively approximate the solutions within the corresponding subsections. Here, we apply a single-layer PIRBN with 61×61 RBF neurons to solve this problem. The centres of the RBF neurons are uniformly distributed within domain $x \in [4.5, 10.5]$, $t \in [4.5, 10.5]$, where the spacing of neurons is 0.05. The initial $b$ is set to 5. **Fig. 22** shows the results from the PIRBN for solving the diffusion equation. As observed from **Fig. 22**(a)-(c), the PIRBN learns the solution well, while the maximum point-wise absolute error is $8\times10^{-3}$. The training process nearly converges after $2\times10^4$ iterations, as shown in **Fig. 22**(d).

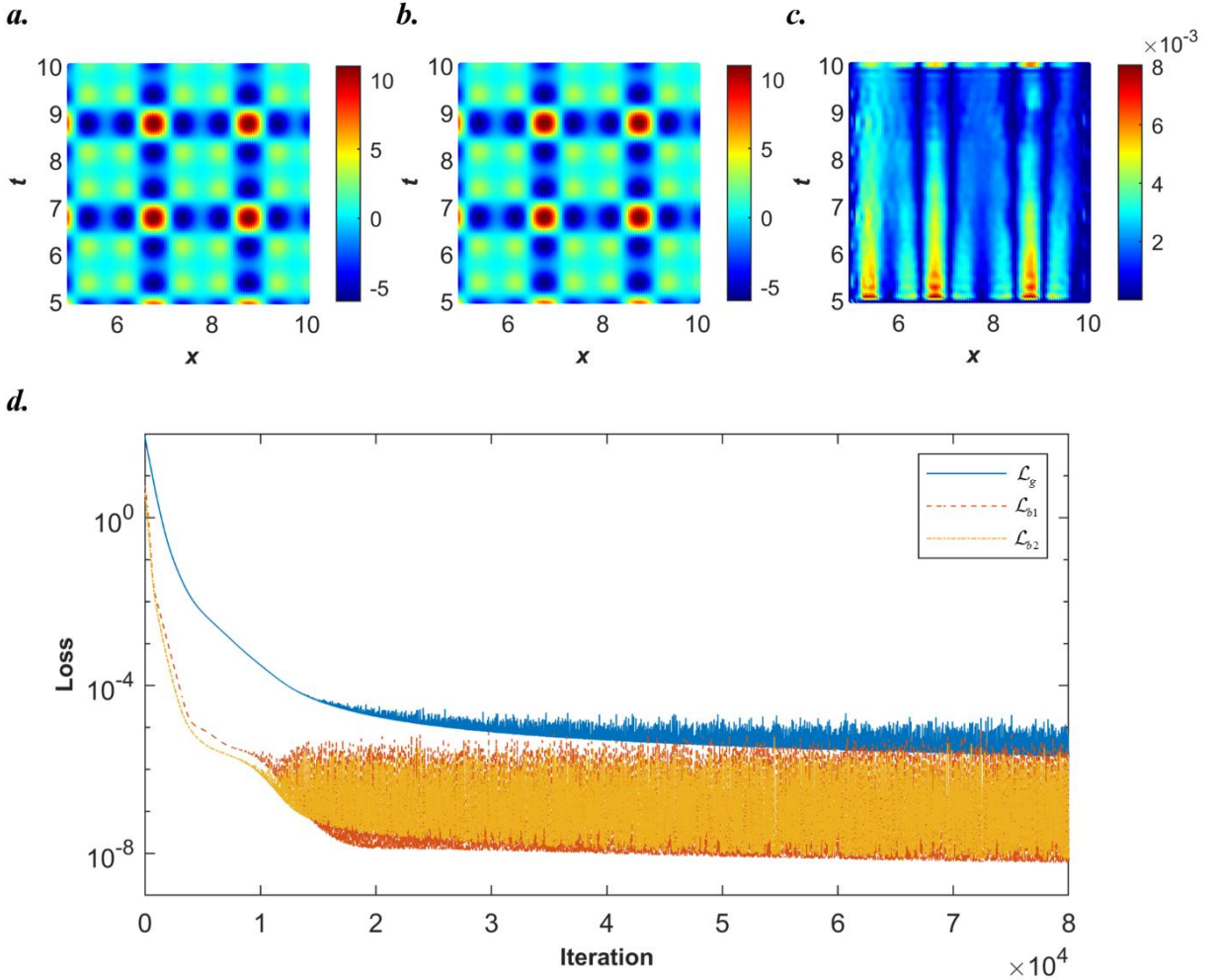

**Fig. 22.** PIRBN results for the diffusion equation. (a) The prediction from the PIRBN. (b) The analytical solution. (c) Point-wise absolute error from the PIRBN. (d) Loss history of the PIRBN during the training process.

### 3.4. Viscoelastic Poiseuille problem

Finally, a viscoelastic Poiseuille flow problem is used to test the performance of PIRBN. Viscoelastic flow is widely observed in nature and is hard to model due to its complexity [63]. Herein, an upper-convected Maxwell (UCM) fluid is used in this problem. The governing equations and the corresponding initial/boundary conditions are given as

$$\begin{cases} \rho \dfrac{\partial}{\partial t} u(y,t) = -f + \dfrac{\partial}{\partial y} \tau_{xy}(y,t), \\ \eta_0 \dfrac{\partial}{\partial y} u(y,t) = \left( \lambda \dfrac{\partial}{\partial t} + 1 \right) \tau_{xy}(y,t), \end{cases} \quad \text{for } t \in [0,4], y \in [-0.5, 0.5]$$
$$u(\pm 0.5, t) = u(y, 0) = 0, \qquad (37)$$
$$\tau_{xy}(y, 0) = 0,$$

where $f = -1.5$ N/m is the constant pressure gradient, $\tau_{xy}$ is the shear stress, $\rho = 1/3$ is the density of the fluid, $\eta_0 = 0.5$ is the viscosity of the fluid, $\lambda = 1/3$ is the relaxation time. The analytical solution is given in Appendix C. In this problem, the evolution contours of physical fields in terms of velocity and shear stress exhibit sharp transition shapes, which are challenging to be captured by PINN. To address the problem, we apply two PIRBNs and each PIRBN contains 26×101 RBF neurons that are uniformly distributed within the domain $t \in$ [-0.2, 4.2], $y \in$ [-0.7, 0.7]. The spacing of RBF neurons is 0.04. The initial $b$ is set to 20.

**Fig. 23** shows the results from the two PIRBNs trained by the Adam optimiser after $10^4$ iterations (learning rate 0.001). For comparison, the predictions by using PINNs are provided in **Fig. 24**. As shown in **Fig. 23**(a)-(d), the evolutions of physics fields in terms of velocity and shear stress are well predicted by the two PIRBNs. The high accuracies of PIRBNs are further demonstrated by the point-wise absolute error contours, which are shown in **Fig. 23**(e) and (f). On the other hand, predictions from the PINNs fail to capture the sharp transition shapes for velocity and shear stress fields, as shown in **Fig. 24**(a)-(d). Furthermore, the point-wise absolute error contours from the PINNs show the large deviations between the PINNs' prediction and the analytical solutions. Finally, by comparing the training histograms shown in **Fig. 23**(g) and **Fig. 24**(g), the PIRBNs achieve lower final losses than the PINNs. The training of the PIRBNs approximately converges after 5000 iterations.

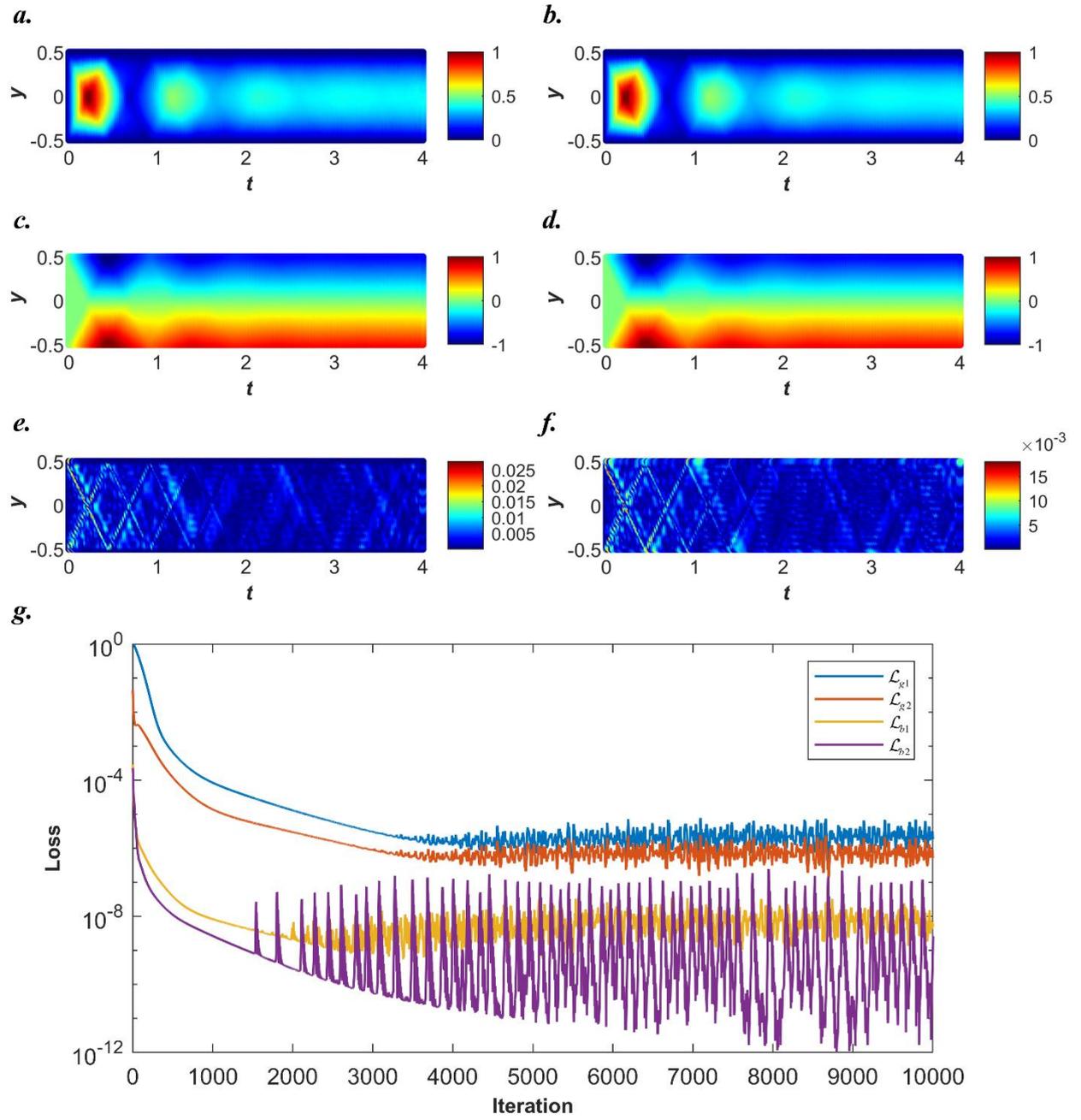

**Fig. 23.** Results from two PIRBNs for the UCM Poiseuille flow problem. (a) The velocity *u* predicted by the PIRBN. (b) The analytical solution of the velocity *u*. (c) The shear stress $\tau_{xy}$ predicted by the PIRBN. (d) The analytical solution of the shear stress $\tau_{xy}$. (e) The point-wise absolute error of the velocity *u* predicted by the PIRBN. (f) The point-wise absolute error of the shear stress $\tau_{xy}$ predicted by the PIRBN. (g) Loss history of the PIRBN during the training process.

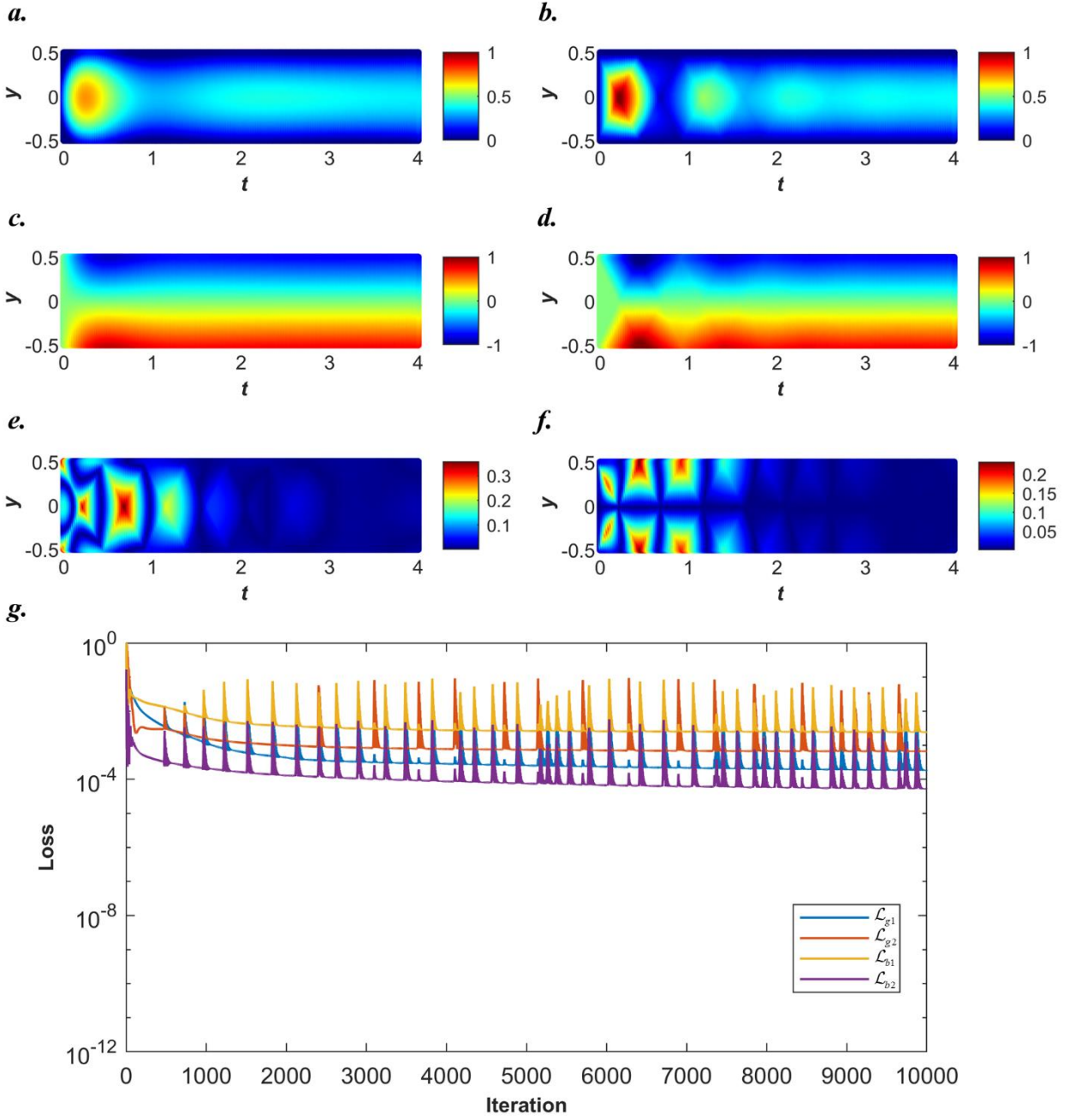

**Fig. 24.** Results from two PINNs for the UCM Poiseuille flow problem. (a) The velocity $u$ predicted by the PINN. (b) The analytical solution of the velocity $u$. (c) The shear stress $\tau_{xy}$ predicted by the PINN. (d) The analytical solution of the shear stress $\tau_{xy}$. (e) The point-wise absolute error of the velocity $u$ predicted by the PINN. (f) The point-wise absolute error of the shear stress $\tau_{xy}$ predicted by the PINN. (g) Loss history of the PINN during the training process.

## 5. Conclusion

In this work, based on observations that PINNs tend to be local approximators after training, we proposed the physics-informed radial basis network (PIRBN) that can effectively solve PDEs suffering from high-frequency features and ill-posed computational domains. Meanwhile, we proved that training a PIRBN can converge to Gaussian Processes with wide-enough neural network width and studied the training dynamics of PIRBN using the NTK theory. Additionally, detailed investigations of the proposed PIRBN regarding initialisation strategies, sizes of sample points and selections of RBF neurons are conducted. In the numerical examples, four challenging PDEs with high-frequency features and ill-posed computational domains are used to test the performances of the proposed PIRBN with respect to the original PINN. Compared to PINN, the proposed PIRBN is more effective than PINN for solving PDEs, which exhibit high-frequency features or suffer from ill-posed computational domains.

Yet, it is important to note that the current PIRBN is limited to a single hidden layer structure, whereas the advantageous performance of DL is enhanced by several hidden layers. Merely adding more hidden layers to the PIRBN may cause the loss of the local approximation property. To take advantage of the power of multiple hidden layers, how to properly integrate more hidden layers to the current PIRBN can be challenging. Besides, since the PIRBN enjoys great local properties along with the training process, specific training strategies can be proposed to boost the training efficiency. These topics will be addressed in our future work.

## Author Contribution

**J. Bai:** Conceptualisation, Methodology, Coding, Formal analysis, Writing-Original draft. **G.-R. Liu:** Advising, Scientific Discussions, Writing-Reviewing and Editing. **A. Gupta:** Writing-Reviewing and Reviewing. **L. Alzubaidi:** Writing-Reviewing and Reviewing. **X.-Q. Feng:** Writing-Reviewing and Reviewing. **Y. Gu:** Conceptualisation, Writing-Reviewing and Editing, Supervision.

## Declaration of Competing Interest

The authors declare no competing interests.

## Acknowledgements

Support from the Australian Research Council research grants (IC190100020 and DP 200102546) is gratefully acknowledged (J. Bai).

# Appendix A. NTK change of PINN with multilayer FNN

PINNs with multiple hidden layers FNN also exhibit the local approximation property during the training process. Herein, consider the PDE problem stated as Eq. (13) when $\mu = 4$, an FNN is constructed with three hidden layers and 61 neurons per layer. 101 sample points are uniformly distributed inside the computational domain. The neural network is trained by Adam optimiser with a learning rate of 0.001.

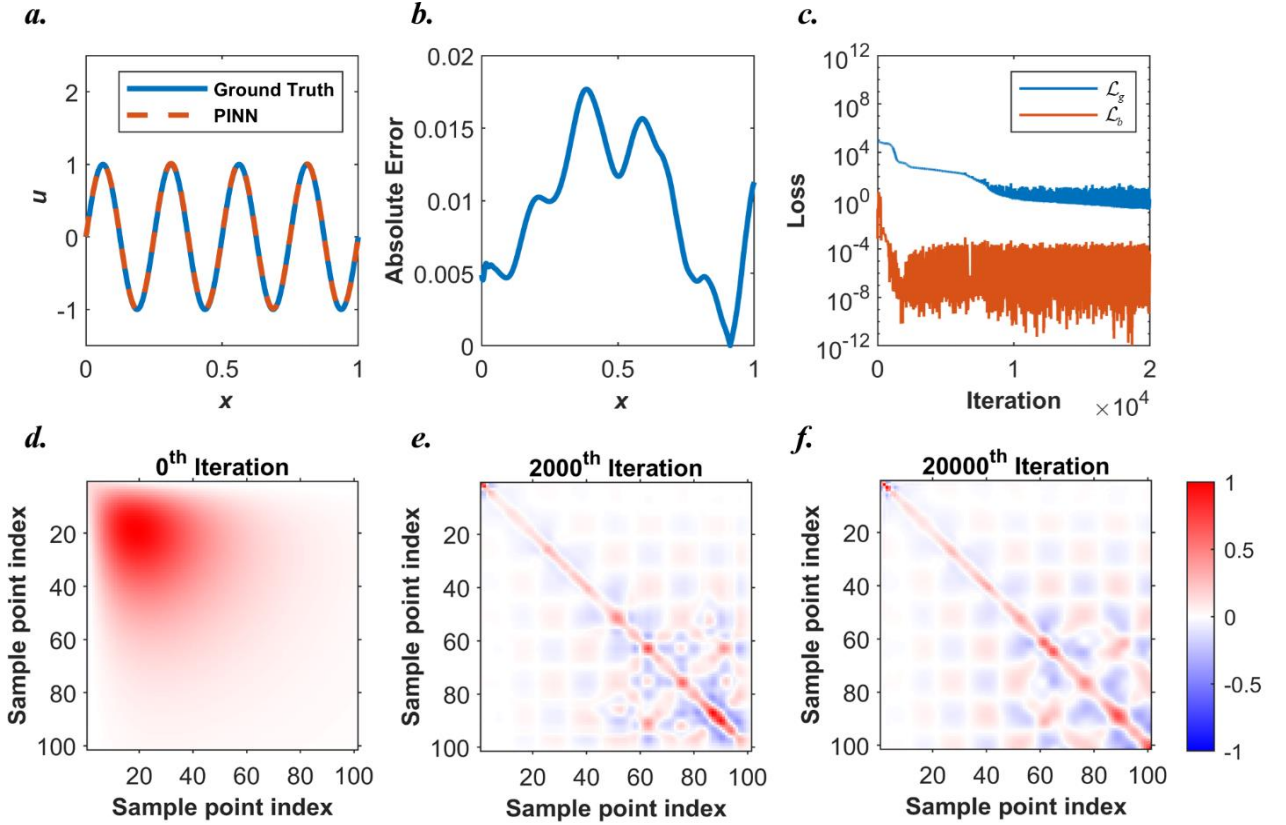

**Fig. A. 1.** Results from a PINN (three hidden layers with 61 neurons per layer) for solving Eq. (13) when $\mu = 4$. (a) Comparisons between the PINN predictions and the ground truth. (b) Point-wise absolute error plot. (c) Loss history of the PINN during the training process. (d) The normalised $\mathbf{K}_g$ at $0^{th}$ iteration. (e) The normalised $\mathbf{K}_g$ at $2000^{th}$ iteration. (f) The normalised $\mathbf{K}_g$ at $20000^{th}$ iteration.

**Fig. A. 1** shows the results from the multilayer FNN after $2\times10^4$ iterations. The prediction from the FNN aligns well with the analytical solution, while the largest point-wise error is less than 0.02. Besides, as observed from **Fig. A. 1**(d)-(f), the NTK of PINN also evolves as a diagonal matrix during training, despite slight oscillations (shadow blue and red colours) in the rest area. In this manner, we conclude that PINN with multiple hidden layers FNN also exhibits the local approximation property during the training process.

# Appendix B. Proofs

## B.1. Proof of theorem 3.1.1

**Proof.** To solve the PDE stated as Eq. (15) by using a PIRBN with the Gaussian function, the second-order partial differential term, $\partial^2 u/\partial x^2$, can be mathematically written as

$$\frac{\partial^2}{\partial x^2} u(x, \boldsymbol{\theta}) = \frac{1}{\sqrt{d}} \sum_{i}^{d} a_i \ddot{\mathcal{G}}_i(x, \boldsymbol{\theta})$$
$$= \frac{1}{\sqrt{d}} \sum_{i}^{d} \left[ -2a_i b_i^2 + 4a_i b_i^4 (x - c_i)^2 \right] e^{-b_i^2 (x - c_i)^2} \quad \text{(B.1)}$$

Recall that all $a_i$ are initialised as i.i.d. random variables satisfy $\mathcal{N}(0,1)$ distribution, $b_i$ and $c_i$ are initialised by given constants. Then, by the central limit theorem, we obtain

$$\mathbb{E}\left[a\ddot{\mathcal{G}}(x)\right] \sim \mathcal{N}(0, \Sigma(x)), \quad \text{(B.2)}$$

where

$$\Sigma(x) = \mathrm{Var}\left[a_i \ddot{\mathcal{G}}(x)\right]. \quad \text{(B.3)}$$

This can be further proved by numerical examples, as shown in **Fig. B. 1**. Note that $e^{-b_i^2(x-c_i)^2} \in (0,1]$,

$$\sup_{d} \mathbb{E}\left[\left|\frac{\partial^2}{\partial x^2} u(x, a, b)\right|^2\right] = \mathbb{E}\left[\frac{1}{d} \sum_{i}^{d} a_i^2 \left(-2b_i^2 + 4b_i^4(x-c_i)^2\right)^2 e^{-2b_i^2(x-c_i)^2}\right]$$
$$\leq \mathbb{E}\left[\frac{1}{d} \sum_{i}^{d} a_i^2 \left(-2b_i^2 + 4b_i^4(x-c_i)^2\right)^2\right] \quad \text{(B.4)}$$
$$= \mathbb{E}\left[a_i^2 \left(-2b_i^2 + 4b_i^4(x-c_i)^2\right)^2\right] < \infty.$$

Thus, we can obtain

$$\Sigma_g(x, x') \triangleq \mathrm{Cov}(\frac{\partial^2 u(x)}{\partial x^2}, \frac{\partial^2 u(x')}{\partial x^2})$$
$$= \mathbb{E}\left[\left(\frac{\partial^2 u(x)}{\partial x^2} - \mathbb{E}\left[\frac{\partial^2 u(x)}{\partial x^2}\right]\right)\left(\frac{\partial^2 u(x')}{\partial x^2} - \mathbb{E}\left[\frac{\partial^2 u(x')}{\partial x^2}\right]\right)\right] \quad \text{(B.5)}$$
$$= \mathbb{E}\left[\frac{1}{d} \sum_{i}^{d} a_i^2 \ddot{\mathcal{G}}(x)\ddot{\mathcal{G}}(x')\right]$$
$$= \mathbb{E}_{a \sim \mathcal{N}(0,1)}\left[a^2 \ddot{\mathcal{G}}(x)\ddot{\mathcal{G}}(x')\right],$$

which completes the proof.

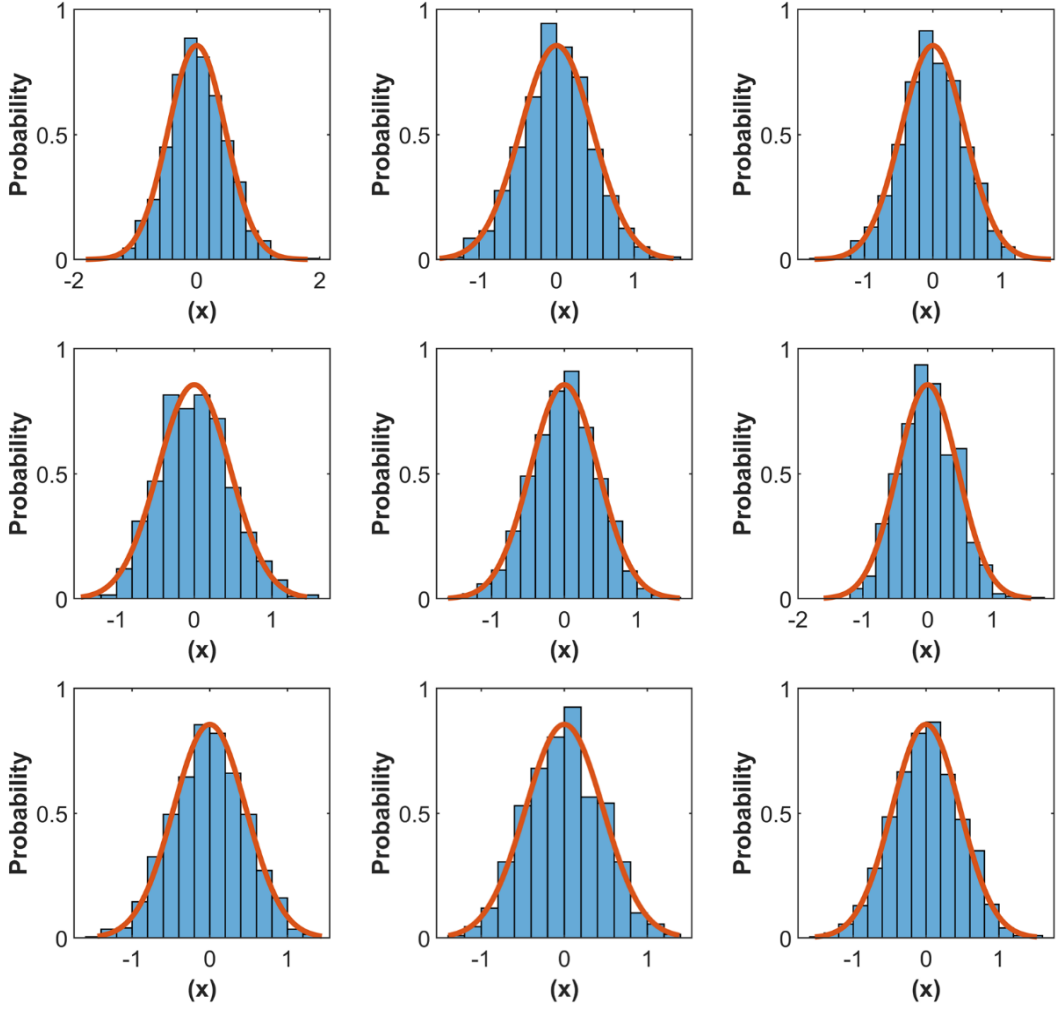

**Fig. B. 1.** The distributions of 1000 values of $a_i \ddot{\vartheta}(x)$ from 9 times random initialisations, where blue columns represent the distribution values within a given range and red lines depict the reference Gaussian distribution. For each initialisation, the distribution of the 1000 values of $a_i \ddot{\vartheta}(x)$ exhibits the Gaussian distribution.

## B.2. Proof of theorem 3.1.3

**Proof.** As defined by Eq. (6), **K**(0) is consist of $\mathbf{K}_{gg}(0)$, $\mathbf{K}_{gb}(0)$ and $\mathbf{K}_{bb}(0)$. Here, we start from $\mathbf{K}_{bb}(0)$. Recalling Eq. (7), for any two given inputs $x$ and $x'$, we can calculate the corresponding element $K_{bb}(0)$ as

$$K_{bb}(0) = \left\langle \frac{\mathrm{dB}\left[u(x;\boldsymbol{\theta}(t))\right]}{\mathrm{d}\boldsymbol{\theta}}, \frac{\mathrm{dB}\left[u(x';\boldsymbol{\theta}(t))\right]}{\mathrm{d}\boldsymbol{\theta}} \right\rangle, \quad (\text{B.6})$$

where $\boldsymbol{\theta} = (a_i, b_i)$, $i = 1, 2, \ldots d$. Given that a PIRBN can be mathematically expressed as

$$u(x;\boldsymbol{\theta}) = \frac{1}{\sqrt{d}} \sum_i^d a_i e^{-b_i^2 (x-c_i)^2}. \quad (\text{B.7})$$

For simplicity, we pre-define

$$\Theta_i = e^{-b_i^2\left[(x-c_i)^2 + (x'-c_i)^2\right]} \in (0,1]. \tag{B.8}$$

Besides, $a_i$, $b_i$ and $c_i$ are independent variables to each other and $\mathbb{E}[a_i]=1$. Therefore, by the central limit theorem, the $K_{bb}(0)$ can be further written as

$$K_{bb}(0) = \mathcal{A}_{bb} + \mathcal{B}_{bb}, \tag{B.9}$$

where

$$\begin{aligned}
\mathcal{A}_{bb} &= \sum_i^d \frac{\partial \mathrm{B}\left[u(x;\boldsymbol{\theta}(t))\right]}{\partial a_i} \frac{\partial \mathrm{B}\left[u(x';\boldsymbol{\theta}(t))\right]}{\partial a_i} \\
&= \frac{1}{d}\sum_i^d \Theta_i \\
&= \mathbb{E}[\Theta], \\
\mathcal{B}_{bb} &= \sum_i^d \frac{\partial \mathrm{B}\left[u(x;\boldsymbol{\theta}(t))\right]}{\partial b_i} \frac{\partial \mathrm{B}\left[u(x';\boldsymbol{\theta}(t))\right]}{\partial b_i} \\
&= -\frac{4}{d}\sum_i^d a_i^2 b_i^2 (x-c_i)^2 (x'-c_i)^2 \Theta_i \\
&= -4\mathbb{E}\left[b^2(x-c)^2(x'-c)^2 \Theta\right],
\end{aligned} \tag{B.10}$$

when $d \to \infty$. Thus, by substituting Eq. (B.10) into Eq. (B.9), we obtain

$$K_{bb}(0) = \mathbb{E}\left[\left(1 + 4b^2(x-c)^2(x'-c)^2\right)\Theta\right]. \tag{B.11}$$

Then, we target on $\mathbf{K}_{gb}(0)$. Again, for any two given inputs $x$ and $x'$, we can calculate the corresponding element $K_{gb}(0)$ as

$$\begin{aligned}
K_{gb}(0) &= \left\langle \frac{\mathrm{dG}\left[u(x;\boldsymbol{\theta}(t))\right]}{\mathrm{d}\boldsymbol{\theta}}, \frac{\mathrm{dB}\left[u(x';\boldsymbol{\theta}(t))\right]}{\mathrm{d}\boldsymbol{\theta}} \right\rangle \\
&= \mathcal{A}_{gb} + \mathcal{B}_{gb},
\end{aligned} \tag{B.12}$$

where

$$\begin{aligned}
\mathcal{A}_{gb} &= \sum_i^d \frac{\partial G[u(x;\theta(t))]}{\partial a_i} \frac{\partial B[u(x';\theta(t))]}{\partial a_i} \\
&= \frac{2}{d}\sum_i^d \left(2b_i^2(x-c_i)^2 - b_i^2\right)\Theta_i \\
&= 2\mathbb{E}\left[\left(2b^2(x-c)^2 - b^2\right)\Theta\right], \\
\mathcal{B}_{gb} &= \sum_i^d \frac{\partial G[u(x;\theta(t))]}{\partial b_i} \frac{\partial B[u(x';\theta(t))]}{\partial b_i} \\
&= \frac{8}{d}\sum_i^d a_i^2 b_i^2 \left(2b_i^4(x-c_i)^4 - 5b_i^2(x-c_i)^2 + 1\right)(x'-c_i)^2 \Theta_i \\
&= 8\mathbb{E}\left[b^2\left(2b^4(x-c)^4 - 5b^2(x-c)^2 + 1\right)(x'-c)^2 \Theta\right],
\end{aligned} \tag{B.13}$$

when $d \to \infty$. Thus, by substituting Eq. (B.10) into Eq. (B.9), we obtain

$$K_{gb}(0) = \mathbb{E}\left[2\left(\left(2b^2(x-c)^2 - b^2\right) + 4b^2\left(2b^4(x-c)^4 - 5b^2(x-c)^2 + 1\right)(x'-c)^2\right)\Theta\right]. \tag{B.14}$$

Finally, we focus on $\mathbf{K}_{gg}(0)$. For any two given inputs $x$ and $x'$, we can calculate the corresponding element $K_{gg}(0)$ as

$$\begin{aligned}
K_{gg}(0) &= \left\langle \frac{dG[u(x;\theta(t))]}{d\theta}, \frac{dG[u(x';\theta(t))]}{d\theta} \right\rangle \\
&= \mathcal{A}_{gg} + \mathcal{B}_{gg},
\end{aligned} \tag{B.15}$$

where

$$\begin{aligned}
\mathcal{A}_{gg} &= \sum_i^d \frac{\partial G[u(x;\theta(t))]}{\partial a_i} \frac{\partial G[u(x';\theta(t))]}{\partial a_i} \\
&= \frac{4}{d}\sum_i^d b_i^4 \left(2b_i^2(x-c_i)^2 - 1\right)\left(2b_i^2(x'-c_i)^2 - 1\right)\Theta_i \\
&= 4\mathbb{E}\left[b^4\left(2b^2(x-c)^2 - 1\right)\left(2b^2(x'-c)^2 - 1\right)\Theta\right], \\
\mathcal{B}_{gg} &= \sum_i^d \frac{\partial G[u(x;\theta(t))]}{\partial b_i} \frac{\partial G[u(x';\theta(t))]}{\partial b_i} \\
&= \frac{16}{d}\sum_i^d a_i^2 b_i^2 \left(2b_i^4(x-c_i)^4 - 5b_i^2(x-c_i)^2 + 1\right)\left(2b_i^4(x'-c_i)^4 - 5b_i^2(x'-c_i)^2 + 1\right)\Theta_i \\
&= 16\mathbb{E}\left[b^2\left(2b^4(x-c)^4 - 5b^2(x-c)^2 + 1\right)\left(2b^4(x'-c)^4 - 5b^2(x'-c)^2 + 1\right)\Theta\right],
\end{aligned} \tag{B.16}$$

when $d \to \infty$. Thus, by substituting Eq. (B.10) into Eq. (B.9), we obtain

$$K_{gg}(0) = \mathbb{E}\Big[ 4\big(b^4\big(2b^2(x-c)^2 - 1\big)\big(2b^2(x'-c)^2 - 1\big) + $$
$$4b^2\big(2b^4(x-c)^4 - 5b^2(x-c)^2 + 1\big)\big(2b^4(x-c)^4 - 5b^2(x'-c)^2 + 1\big)\big)\Theta\Big]. \quad (B.17)$$

Since $b_i$, $c_i$ are initialised with given values, $K_{bb}(0)$, $K_{gb}(0)$ and $K_{gg}(0)$ are convergence to determinate values, which complete the proof.

**B.3. Proof of theorem 3.1.4**

**Proof.** The NTK of the PIRBN can be written as

$$\mathbf{K}(t) = \mathbf{J}(t)\mathbf{J}^{\mathrm{T}}(t), \quad (B.18)$$

where $\mathbf{J}(t)$ is the Jacobian matrix

$$\mathbf{J}(t) = \begin{bmatrix} \mathbf{J}_g(t) \\ \mathbf{J}_b(t) \end{bmatrix} = \begin{bmatrix} \dfrac{\mathrm{dG}\big[u\big(x_i^g;\boldsymbol{\theta}(t)\big)\big]}{\mathrm{d}\boldsymbol{\theta}} \\ \dfrac{\mathrm{dB}\big[u\big(x_i^b;\boldsymbol{\theta}(t)\big)\big]}{\mathrm{d}\boldsymbol{\theta}} \end{bmatrix}. \quad (B.19)$$

By substituting Eq. (B.19) into Eq. (23) and the Cauchy-Schwartz inequality, we have

$$\|\mathbf{K}(t) - \mathbf{K}(0)\|_2 = \|\mathbf{J}(t)\mathbf{J}^T(t) - \mathbf{J}(0)\mathbf{J}^T(0)\|_2$$
$$\leq \|\mathbf{J}(t)\|_2 \|\mathbf{J}(t) - \mathbf{J}(0)\|_2 + \|\mathbf{J}(t) - \mathbf{J}(0)\|_2 \|\mathbf{J}(0)\|_2 \quad (B.20)$$

Here, we start from $\|\mathbf{J}(t)\|_2$ first. Recalling Eq.(7), for any given inputs $x$, the two components of $\mathbf{J}(t)$ can be calculated as

$$\mathbf{J}_g(t) = \begin{bmatrix} \dfrac{\partial \mathrm{G}\big[u(x;\boldsymbol{\theta}(t))\big]}{\partial a_i} & \dfrac{\partial \mathrm{G}\big[u(x;\boldsymbol{\theta}(t))\big]}{\partial b_i} \end{bmatrix}$$
$$= \begin{bmatrix} \dfrac{2}{\sqrt{d}}\big[2b_i^4(x-c_i)^2 - b_i^2\big]e^{-b_i^2(x-c_i)^2} \\ \dfrac{4a_i b_i}{\sqrt{d}}\big[5b_i^2(x-c_i)^2 - 2b_i^4(x-c_i)^4 - 1\big]e^{-b_i^2(x-c_i)^2} \end{bmatrix}^\mathrm{T},$$
$$\mathbf{J}_b(t) = \begin{bmatrix} \dfrac{\partial \mathrm{B}\big[u(x;\boldsymbol{\theta}(t))\big]}{\partial a_i} & \dfrac{\partial \mathrm{B}\big[u(x;\boldsymbol{\theta}(t))\big]}{\partial b_i} \end{bmatrix} \quad (B.21)$$
$$= \begin{bmatrix} \dfrac{1}{\sqrt{d}}e^{b_i^2(x-c_i)^2} & \dfrac{2a_i b_i}{\sqrt{d}}(x-c_i)^2 e^{b_i^2(x-c_i)^2} \end{bmatrix}$$

Since $e^{-b_i^2(x-c_i)^2}$ and $a_i(t)$ are all bounded during the training process, $J_g(t)$ and $J_b(t)$ are bounded during the training process. Thus, $\|\mathbf{J}(t)\|_2$ also is bounded during the training process.

Then, we focus on $\|\mathbf{J}(t) - \mathbf{J}(0)\|_2$. For any given inputs $x$, the two components of $\|\mathbf{J}(t) - \mathbf{J}(0)\|_2$ can be written as

$$\|\mathbf{J}_b(t) - \mathbf{J}_b(0)\|_2 = \left\| \frac{d B[u(x;\boldsymbol{\theta}(t))]}{d\boldsymbol{\theta}} - \frac{d B[u(x;\boldsymbol{\theta}(0))]}{d\boldsymbol{\theta}} \right\|_2,$$
$$\|\mathbf{J}_g(t) - \mathbf{J}_g(0)\|_2 = \left\| \frac{d G[u(x;\boldsymbol{\theta}(t))]}{d\boldsymbol{\theta}} - \frac{d G[u(x;\boldsymbol{\theta}(0))]}{d\boldsymbol{\theta}} \right\|_2$$
(B.22)

For sack of simplicity, we pre-define following terms

$$\begin{aligned}
\varepsilon_i &= (x - c_i)^2 \\
A_i(t) &= e^{-b_i^2(t)\varepsilon_i}, \\
B_i(t) &= a_i(t) b_i(t) \varepsilon_i A_i(t), \\
C_i(t) &= 2\left[ 2 b_i^4(t)\varepsilon_i - b_i^2(t) \right] A_i(t), \\
D_i(t) &= 4 a_i(t) b_i(t) \left[ 5 b_i^2(t)\varepsilon_i - 2 b_i^4(t)\varepsilon_i^2 - 1 \right] A_i(t).
\end{aligned}$$
(B.23)

Thus, for $\|\mathbf{J}_b(t) - \mathbf{J}_b(0)\|_2$ we can obtain

$$\left\| \frac{d B[u(x;\boldsymbol{\theta}(t))]}{d\boldsymbol{\theta}(t)} - \frac{d B[u(x;\boldsymbol{\theta}(0))]}{d\boldsymbol{\theta}(0)} \right\|_2$$
$$\leq \frac{2}{\sqrt{d}} \|\mathbf{A}(t) - \mathbf{A}(0)\|_2 + \frac{2}{\sqrt{d}} \|\mathbf{B}(t) - \mathbf{B}(0)\|_2,$$

$$\left\| \frac{d G[u(x;\boldsymbol{\theta}(t))]}{d\boldsymbol{\theta}(t)} - \frac{d G[u(x;\boldsymbol{\theta}(0))]}{d\boldsymbol{\theta}(0)} \right\|_2$$
$$\leq \frac{1}{\sqrt{d}} \|\mathbf{C}(t) - \mathbf{C}(0)\|_2 + \frac{1}{\sqrt{d}} \|\mathbf{D}(t) - \mathbf{D}(0)\|_2.$$
(B.24)

We first focus on $\frac{2}{\sqrt{d}} \|\mathbf{A}(t) - \mathbf{A}(0)\|_2$. Note that the evolution of $\boldsymbol{\theta}(t)$ between $t_i$ and $t_{i+1}$ iterations using gradient descendent algorithms can be written as

$$\frac{\boldsymbol{\theta}(t_{i+1}) - \boldsymbol{\theta}(t_i)}{\eta} = -\nabla \mathcal{L}(\boldsymbol{\theta}),$$
(B.25)

where $\eta$ is the learning rate. With infinitesimally small learning rate, we can obtain the gradient flow as

$$\frac{d\boldsymbol{\theta}}{dt} = \frac{d\boldsymbol{\theta}}{d\mathbf{A}} \frac{d\mathbf{A}}{dt} = -\nabla \mathcal{L}(\boldsymbol{\theta}). \tag{B.26}$$

By Eq. (B.26), we can obtain

$$\begin{aligned}
&\frac{2}{\sqrt{d}} \left\| \mathbf{A}(t) - \mathbf{A}(0) \right\|_2, \\
&= \frac{2}{\sqrt{d}} \left\| \int_0^t \frac{d\mathbf{A}(\sigma)}{d\sigma} d\sigma \right\|_2 \\
&= \frac{2}{\sqrt{d}} \left\| \int_0^t \frac{d\mathbf{A}(\sigma)}{d\boldsymbol{\theta}} \frac{d\mathcal{L}(\boldsymbol{\theta}(\sigma))}{d\boldsymbol{\theta}} d\sigma \right\|_2 \\
&\leq \frac{2}{\sqrt{d}} \int_0^t \left\| \frac{d\mathbf{A}(\sigma)}{d\boldsymbol{\theta}} \frac{d\mathcal{L}(\boldsymbol{\theta}(\sigma))}{d\boldsymbol{\theta}} \right\|_2 d\sigma, \\
&= \frac{2}{\sqrt{d}} \int_0^t \sqrt{\sum_i^d \left( \frac{dA_i(\sigma)}{db_i} \frac{\partial \mathcal{L}(\boldsymbol{\theta}(\sigma))}{\partial b_i} \right)^2} d\sigma,
\end{aligned} \tag{B.27}$$

where

$$\begin{aligned}
\frac{dA_i(\sigma)}{db_i} &= -2b_i(\sigma)\varepsilon_i e^{-b_i^2(\sigma)\varepsilon_i}, \\
\frac{\partial \mathcal{L}(\boldsymbol{\theta})}{\partial b_i} &= \sum_j^{n_g} \left( G[u(x_j^g, \boldsymbol{\theta})] - g(x_j^g) \right) \frac{\partial G[u(x_j^g, \boldsymbol{\theta})]}{\partial b_i} + \\
&\quad \sum_j^{n_b} \left( B[u(x_j^b, \boldsymbol{\theta})] - b(x_j^b) \right) \frac{\partial B[u(x_j^b, \boldsymbol{\theta})]}{\partial b_i}.
\end{aligned} \tag{B.28}$$

Thus,

$$\begin{aligned}
&\frac{2}{\sqrt{d}} \int_0^t \sqrt{\sum_i^d \left( \frac{dA_i(\sigma)}{db_i} \frac{\partial \mathcal{L}(\boldsymbol{\theta}(\sigma))}{\partial b_i} \right)^2} d\sigma, \\
&\leq \frac{2}{\sqrt{d}} \int_0^t \left\| \frac{d\mathbf{A}(\sigma)}{d\mathbf{b}} \right\|_\infty \sum_i^d \sqrt{\left( \frac{\partial \mathcal{L}(\boldsymbol{\theta}(\sigma))}{\partial b_i} \right)^2} d\sigma \\
&\leq H + I,
\end{aligned} \tag{B.29}$$

where

$$\begin{aligned}
H &= \frac{2}{\sqrt{d}} \int_0^t \left\| \frac{d\mathbf{A}(\sigma)}{d\mathbf{b}} \right\|_\infty \left\| \frac{dG[u(x_j^g)]}{d\mathbf{b}} \right\|_\infty \sqrt{\sum_i^d \left( \sum_j^{n_g} \left( G[u(x_j^g)] - g(x_j^g) \right) \right)^2} d\sigma \\
I &= \frac{2}{\sqrt{d}} \int_0^t \left\| \frac{d\mathbf{A}(\sigma)}{d\mathbf{b}} \right\|_\infty \left\| \frac{dB[u(x_j^b)]}{d\mathbf{b}} \right\|_\infty \sqrt{\sum_i^d \left( \sum_j^{n_b} \left( B[u(x_j^b)] - b(x_j^b) \right) \right)^2} d\sigma
\end{aligned} \tag{B.30}$$

Note that [50]

$$\left\|\frac{d\mathbf{A}(\sigma)}{d\mathbf{b}}\right\|_{\infty} = O\left(\frac{1}{\sqrt{d}}\right),$$
$$\left\|\frac{dG[u(x_j^g)]}{d\mathbf{b}}\right\|_{\infty} = O\left(\frac{1}{\sqrt{d}}\right), \quad \text{(B.31)}$$
$$\left\|\frac{dB[u(x_j^b)]}{d\mathbf{b}}\right\|_{\infty} = O\left(\frac{1}{\sqrt{d}}\right).$$

Hence,

$$\lim_{d\to\infty} \sup_{t\in[0,T]} H = O\left(\frac{2\mathcal{C}}{d}\right) = 0,$$
$$\lim_{d\to\infty} \sup_{t\in[0,T]} I = O\left(\frac{2\mathcal{C}}{d}\right) = 0, \quad \text{(B.32)}$$

We can conclude that

$$\lim_{d\to\infty} \sup_{t\in[0,T]} \frac{2}{\sqrt{d}} \|\mathbf{A}(t) - \mathbf{A}(0)\|_2 = 0. \quad \text{(B.33)}$$

Similarly, we can prove that

$$\lim_{d\to\infty} \sup_{t\in[0,T]} \frac{2}{\sqrt{d}} \|\mathbf{B}(t) - \mathbf{B}(0)\|_2 = 0,$$
$$\lim_{d\to\infty} \sup_{t\in[0,T]} \frac{2}{\sqrt{d}} \|\mathbf{C}(t) - \mathbf{C}(0)\|_2 = 0, \quad \text{(B.34)}$$
$$\lim_{d\to\infty} \sup_{t\in[0,T]} \frac{2}{\sqrt{d}} \|\mathbf{D}(t) - \mathbf{D}(0)\|_2 = 0.$$

By Eq. (B.33) and Eq. (B.34), we obtain that

$$\lim_{d\to\infty} \sup_{t\in[0,T]} \left(\frac{2}{\sqrt{d}} \|\mathbf{A}(t) - \mathbf{A}(0)\|_2 + \frac{2}{\sqrt{d}} \|\mathbf{B}(t) - \mathbf{B}(0)\|_2\right) \to 0,$$
$$\lim_{d\to\infty} \sup_{t\in[0,T]} \left(\frac{1}{\sqrt{d}} \|\mathbf{C}(t) - \mathbf{C}(0)\|_2 + \frac{1}{\sqrt{d}} \|\mathbf{D}(t) - \mathbf{D}(0)\|_2\right) \to 0. \quad \text{(B.35)}$$

Therefore, we have

$$\lim_{d\to\infty} \sup_{t\in[0,T]} \|\mathbf{J}(t) - \mathbf{J}(0)\|_2 \to 0. \quad \text{(B.36)}$$

And obviously, one can conclude that

$$\lim_{d \to \infty} \sup_{t \in [0,T]} \|\mathbf{K}(t) - \mathbf{K}(0)\|_2 \to 0. \tag{B.37}$$

The above contents complete the proof.

## Appendix C. The analytical solution of Poiseuille flow with the UCM fluid

The analytical solution of the Poiseuille flow problem with the UCM fluid is given as [64]

$$\begin{aligned} u(y,t) &= u_{\max} U(y,t), \\ \tau_{xy}(y,t) &= u_{\max} T_{xy}(y,t) \end{aligned} \tag{C.1}$$

where $u_{\max} = 4h^2 f/(12\eta_0)$ is the mean streamwise velocity. $U(y, t)$ and $T_{xy}(y, t)$ are given as

$$\begin{aligned} U(y,t) &= 1.5(1-4y^2) - 48 \sum_{i=1}^{\infty} \frac{\sin(0.5 N_i(2y+1))}{N_i^3} e^{-0.5t/\lambda} G_i(t), \\ T_{xy}(y,t) &= -6Y + 48 \sum_{i=1}^{\infty} \frac{\cos(0.5 N_i(2y+1))}{N_i} e^{-0.5t/\lambda} Q_i(t), \end{aligned} \tag{C.2}$$

where

$$\begin{aligned} G_i(t) &= \cosh\left(\frac{\beta_i t}{2\lambda}\right) + \frac{\gamma_i}{\beta_i} \sinh(\frac{\beta_i t}{2\lambda}), \\ Q_i(t) &= -\frac{\alpha}{E N_i^3} G_i(t) + \frac{2}{\lambda E N_i^3} \frac{\partial}{\partial t} G_i(t), \end{aligned} \tag{C.3}$$

and

$$\begin{aligned} \beta_i &= \sqrt{E N_i^2 - 1}, \gamma_i = 1 - 0.5 E N_i^2, \\ N_i &= (2i-1)\pi, E = \frac{\lambda \eta_0}{\rho h^2}. \end{aligned} \tag{C.4}$$

## Reference

[1] C. Beck, M. Hutzenthaler, A. Jentzen, B.J.a.p.a. Kuckuck, An overview on deep learning-based approximation methods for partial differential equations, (2020).

[2] G.E. Karniadakis, I.G. Kevrekidis, L. Lu, P. Perdikaris, S. Wang, L. Yang, Physics-informed machine learning, Nature Reviews Physics, 3 (2021) 422-440.

[3] J. Bai, L. Alzubaidi, Q. Wang, E. Kuhl, M. Bennamoun, Y. Gu, Utilising physics-guided deep learning to overcome data scarcity, Arxiv preprint arXiv:2211.15664, (2022).

[4] E. Samaniego, C. Anitescu, S. Goswami, V.M. Nguyen-Thanh, H. Guo, K. Hamdia, X. Zhuang, T. Rabczuk, An energy approach to the solution of partial differential equations in

computational mechanics via machine learning: Concepts, implementation and applications, Computer Methods in Applied Mechanics and Engineering, 362 (2020).

[5] V.M. Nguyen-Thanh, X. Zhuang, T. Rabczuk, A deep energy method for finite deformation hyperelasticity, European Journal of Mechanics - A/Solids, 80 (2020).

[6] W. Li, M.Z. Bazant, J. Zhu, A physics-guided neural network framework for elastic plates: Comparison of governing equations-based and energy-based approaches, Computer Methods in Applied Mechanics and Engineering, 383 (2021).

[7] E. Haghighat, M. Raissi, A. Moure, H. Gomez, R. Juanes, A physics-informed deep learning framework for inversion and surrogate modeling in solid mechanics, Computer Methods in Applied Mechanics and Engineering, 379 (2021).

[8] X. Zhuang, H. Guo, N. Alajlan, H. Zhu, T. Rabczuk, Deep autoencoder based energy method for the bending, vibration, and buckling analysis of Kirchhoff plates with transfer learning, European Journal of Mechanics - A/Solids, 87 (2021).

[9] A. Henkes, H. Wessels, R. Mahnken, Physics informed neural networks for continuum micromechanics, Computer Methods in Applied Mechanics and Engineering, 393 (2022).

[10] S. Rezaei, A. Harandi, A. Moeineddin, B.-X. Xu, S. Reese, A mixed formulation for physics-informed neural networks as a potential solver for engineering problems in heterogeneous domains: Comparison with finite element method, Computer Methods in Applied Mechanics and Engineering, 401 (2022).

[11] J. Bai, Y. Zhou, Y. Ma, H. Jeong, H. Zhan, C. Rathnayaka, E. Sauret, Y. Gu, A general Neural Particle Method for hydrodynamics modeling, Computer Methods in Applied Mechanics and Engineering, 393 (2022).

[12] H. Wessels, C. Weißenfels, P. Wriggers, The neural particle method – An updated Lagrangian physics informed neural network for computational fluid dynamics, Computer Methods in Applied Mechanics and Engineering, 368 (2020).

[13] M. Raissi, A. Yazdani, G.E. Karniadakis, Hidden fluid mechanics: Learning velocity and pressure fields from flow visualizations, Science, 367 (2020) 1026-1030.

[14] S. Cai, Z. Wang, F. Fuest, Y.J. Jeon, C. Gray, G.E. Karniadakis, Flow over an espresso cup: inferring 3-D velocity and pressure fields from tomographic background oriented Schlieren via physics-informed neural networks, Journal of Fluid Mechanics, 915 (2021).

[15] Z. Li, J. Bai, H. Ouyang, S. Martelli, M. Tang, H. Wei, P. Liu, W.-R. Han, Y. Gu, Physics-informed neutral network for friction-involved nonsmooth dynamics problems, arXiv preprint arXiv:.02542, (2023).

[16] A. Ghaderi, V. Morovati, Y. Chen, R. Dargazany, A physics-informed multi-agents model to predict thermo-oxidative/hydrolytic aging of elastomers, International Journal of Mechanical Sciences, 223 (2022).

[17] J. Wu, L. Bai, J. Huang, L. Ma, J. Liu, S. Liu, Accurate force field of two-dimensional ferroelectrics from deep learning, Physical Review B, 104 (2021).

[18] W. Li, J. Zhang, F. Ringbeck, D. Jöst, L. Zhang, Z. Wei, D.U. Sauer, Physics-informed neural networks for electrode-level state estimation in lithium-ion batteries, Journal of Power Sources, 506 (2021).

[19] Y. Chen, L. Lu, G.E. Karniadakis, L. Dal Negro, Physics-informed neural networks for inverse problems in nano-optics and metamaterials, Opt Express, 28 (2020) 11618-11633.45

[20] L. Zhang, J. Han, H. Wang, R. Car, W. E, Deep Potential Molecular Dynamics: A Scalable Model with the Accuracy of Quantum Mechanics, Phys Rev Lett, 120 (2018) 143001.

[21] H. Wang, L. Zhang, J. Han, W. E, DeePMD-kit: A deep learning package for many-body potential energy representation and molecular dynamics, Computer Physics Communications, 228 (2018) 178-184.

[22] J. Tian, R. Xiong, J. Lu, C. Chen, W. Shen, Battery state-of-charge estimation amid dynamic usage with physics-informed deep learning, Energy Storage Materials, 50 (2022) 718-729.

[23] M. Yin, X. Zheng, J.D. Humphrey, G. Em Karniadakis, Non-invasive Inference of Thrombus Material Properties with Physics-Informed Neural Networks, Comput Methods Appl Mech Eng, 375 (2021).

[24] G. Kissas, Y. Yang, E. Hwuang, W.R. Witschey, J.A. Detre, P. Perdikaris, Machine learning in cardiovascular flows modeling: Predicting arterial blood pressure from non-invasive 4D flow MRI data using physics-informed neural networks, Computer Methods in Applied Mechanics and Engineering, 358 (2020).

[25] K. Linka, A. Schäfer, X. Meng, Z. Zou, G.E. Karniadakis, E. Kuhl, Bayesian Physics Informed Neural Networks for real-world nonlinear dynamical systems, Computer Methods in Applied Mechanics and Engineering, (2022).

[26] D.R. Rutkowski, A. Roldan-Alzate, K.M. Johnson, Enhancement of cerebrovascular 4D flow MRI velocity fields using machine learning and computational fluid dynamics simulation data, Sci Rep, 11 (2021) 10240.

[27] E. Kharazmi, M. Cai, X. Zheng, Z. Zhang, G. Lin, G.E. Karniadakis, Identifiability and predictability of integer- and fractional-order epidemiological models using physics-informed neural networks, Nature Computational Science, 1 (2021) 744-753.

[28] M.P.T. Kaandorp, S. Barbieri, R. Klaassen, H.W.M. van Laarhoven, H. Crezee, P.T. While, A.J. Nederveen, O.J. Gurney-Champion, Improved unsupervised physics-informed deep learning for intravoxel incoherent motion modeling and evaluation in pancreatic cancer patients, Magn Reson Med, 86 (2021) 2250-2265.

[29] S. Cai, H. Li, F. Zheng, F. Kong, M. Dao, G.E. Karniadakis, S. Suresh, Artificial intelligence velocimetry and microaneurysm-on-a-chip for three-dimensional analysis of blood flow in physiology and disease, Proc Natl Acad Sci U S A, 118 (2021).

[30] S. Buoso, T. Joyce, S. Kozerke, Personalising left-ventricular biophysical models of the heart using parametric physics-informed neural networks, Med Image Anal, 71 (2021) 102066.

[31] J.H. Lagergren, J.T. Nardini, R.E. Baker, M.J. Simpson, K.B. Flores, Biologically-informed neural networks guide mechanistic modeling from sparse experimental data, PLoS Comput Biol, 16 (2020) e1008462.

[32] F. Sahli Costabal, Y. Yang, P. Perdikaris, D.E. Hurtado, E. Kuhl, Physics-Informed Neural Networks for Cardiac Activation Mapping, Frontiers in Physics, 8 (2020).

[33] A. Rodríguez, J. Cui, N. Ramakrishnan, B. Adhikari, B.A. Prakash, EINNs: Epidemiologically-Informed Neural Networks, arXiv preprint arXiv:.10446, (2022).

[34] M.A. Bhouri, F.S. Costabal, H. Wang, K. Linka, M. Peirlinck, E. Kuhl, P. Perdikaris, COVID-19 dynamics across the US: A deep learning study of human mobility and social behavior, Computer Methods in Applied Mechanics and Engineering, 382 (2021).

[35] S. Wang, Y. Teng, P. Perdikaris, Understanding and mitigating gradient pathologies in physics-informed neural networks, arXiv preprint arXiv:.04536, (2020).

[36] H. Jeong, J. Bai, C.P. Batuwatta-Gamage, C. Rathnayaka, Y. Zhou, Y. Gu, A Physics-Informed Neural Network-based Topology Optimization (PINNTO) framework for structural optimization, Engineering Structures, 278 (2023).

[37] E. Kharazmi, Z. Zhang, G.E.M. Karniadakis, hp-VPINNs: Variational physics-informed neural networks with domain decomposition, Computer Methods in Applied Mechanics and Engineering, 374 (2021).

[38] E. Haghighat, A.C. Bekar, E. Madenci, R. Juanes, A nonlocal physics-informed deep learning framework using the peridynamic differential operator, Computer Methods in Applied Mechanics and Engineering, 385 (2021).

[39] J. Yu, L. Lu, X. Meng, G.E. Karniadakis, Gradient-enhanced physics-informed neural networks for forward and inverse PDE problems, Computer Methods in Applied Mechanics and Engineering, 393 (2022).

[40] J.N. Fuhg, N. Bouklas, The mixed Deep Energy Method for resolving concentration features in finite strain hyperelasticity, Journal of Computational Physics, (2021).

[41] J. Bai, T. Rabczuk, A. Gupta, L. Alzubaidi, Y. Gu, A physics-informed neural network technique based on a modified loss function for computational 2D and 3D solid mechanics, Computational Mechanics, (2022).

[42] A.D. Jagtap, E. Kharazmi, G.E. Karniadakis, Conservative physics-informed neural networks on discrete domains for conservation laws: Applications to forward and inverse problems, Computer Methods in Applied Mechanics and Engineering, 365 (2020).

[43] S. Dong, Z. Li, Local extreme learning machines and domain decomposition for solving linear and nonlinear partial differential equations, Computer Methods in Applied Mechanics and Engineering, 387 (2021).

[44] K. Shukla, A.D. Jagtap, G.E. Karniadakis, Parallel physics-informed neural networks via domain decomposition, Journal of Computational Physics, 447 (2021).

[45] V.M. Nguyen-Thanh, C. Anitescu, N. Alajlan, T. Rabczuk, X. Zhuang, Parametric deep energy approach for elasticity accounting for strain gradient effects, Computer Methods in Applied Mechanics and Engineering, 386 (2021).

[46] A. Jacot, F. Gabriel, C. Hongler, Neural tangent kernel: Convergence and generalization in neural networks, Advances in neural information processing systems, 31 (2018).

[47] Y. Chen, B. Hosseini, H. Owhadi, A.M. Stuart, Solving and learning nonlinear PDEs with Gaussian processes, Journal of Computational Physics, 447 (2021) 110668.

[48] S.S. Du, X. Zhai, B. Poczos, A. Singh, Gradient descent provably optimizes over-parameterized neural networks, arXiv preprint arXiv:.02054, (2018).

[49] S.S. Du, J. Lee, H. Li, L. Wang, X. Zhai, Gradient Descent Finds Global Minima of Deep Neural Networks, in: Proc. Proceedings of the 36th International Conference on Machine Learning, PMLR, Proceedings of Machine Learning Research, 2019, 1675--1685.

[50] S. Wang, X. Yu, P. Perdikaris, When and why PINNs fail to train: A neural tangent kernel perspective, Journal of Computational Physics, (2021).

[51] S. Wang, H. Wang, P. Perdikaris, On the eigenvector bias of Fourier feature networks: From regression to solving multi-scale PDEs with physics-informed neural networks, Computer Methods in Applied Mechanics and Engineering, 384 (2021).

[52] S. Wang, P. Perdikaris, Long-time integration of parametric evolution equations with physics-informed DeepONets, Journal of Computational Physics, (2022).

[53] Y. Li, S. Guo, Z. Gan, Empirical prior based probabilistic inference neural network for policy learning, Information Sciences, 615 (2022) 678-699.

[54] C. Xu, B.T. Cao, Y. Yuan, G. Meschke, Transfer learning based physics-informed neural networks for solving inverse problems in engineering structures under different loading scenarios, Computer Methods in Applied Mechanics and Engineering, 405 (2023).

[55] M. Raissi, P. Perdikaris, G.E. Karniadakis, Physics-informed neural networks: A deep learning framework for solving forward and inverse problems involving nonlinear partial differential equations, Journal of Computational Physics, 378 (2019) 686-707.

[56] Z. Lin, V. Sekar, G. Fanti, Why spectral normalization stabilizes gans: Analysis and improvements, Advances in neural information processing systems, 34 (2021) 9625-9638.

[57] Y. Bengio, I. Goodfellow, A. Courville, Deep learning, MIT press Massachusetts, USA:, 2017.

[58] D.S. Broomhead, D. Lowe, Radial basis functions, multi-variable functional interpolation and adaptive networks, in, Royal Signals and Radar Establishment Malvern (United Kingdom), 1988.

[59] G.-R. Liu, Y. Gu, An introduction to meshfree methods and their programming, Springer Science & Business Media, 2005.

[60] B. Bordelon, A. Canatar, C. Pehlevan, Spectrum dependent learning curves in kernel regression and wide neural networks, in: International Conference on Machine Learning, PMLR, 2020, pp. 1024-1034.

[61] C. Zhang, M. Rezavand, Y. Zhu, Y. Yu, D. Wu, W. Zhang, J. Wang, X.J.C.P.C. Hu, SPHinXsys: An open-source multi-physics and multi-resolution library based on smoothed particle hydrodynamics, 267 (2021) 108066.

[62] C.P. Batuwatta-Gamage, C.M. Rathnayaka, H.C.P. Karunasena, W.D.C.C. Wijerathne, H. Jeong, Z.G. Welsh, M.A. Karim, Y.T. Gu, A physics-informed neural network-based surrogate framework to predict moisture concentration and shrinkage of a plant cell during drying, Journal of Food Engineering, 332 (2022).

[63] M.A. Alves, P.J. Oliveira, F.T. Pinho, Numerical Methods for Viscoelastic Fluid Flows, Annual Review of Fluid Mechanics, 53 (2021) 509-541.

[64] S.C. Xue, R.I. Tanner, N. Phan-Thien, Numerical modelling of transient viscoelastic flows, Journal of Non-Newtonian Fluid Mechanics, 123 (2004) 33-58.